\documentclass{article} 
\usepackage{iclr2027_conference,times}

\usepackage{amsmath,amsfonts,bm}

\def\eqref#1{equation~\ref{#1}}

\def\1{\bm{1}}

\DeclareMathAlphabet{\mathsfit}{\encodingdefault}{\sfdefault}{m}{sl}
\SetMathAlphabet{\mathsfit}{bold}{\encodingdefault}{\sfdefault}{bx}{n}

\usepackage{hyperref}
\usepackage{url}
\usepackage{svg}
\usepackage{booktabs}
\usepackage{multirow}
\usepackage{graphicx}
\usepackage{float}
\usepackage{placeins}
\usepackage{tabularx}
\usepackage{amssymb}
\usepackage{wrapfig}
\usepackage{array,booktabs,longtable}
\usepackage{listings}
\usepackage{xcolor}
\usepackage{algorithm}
\usepackage{algpseudocode}
\usepackage{wrapfig}
\usepackage{graphicx}
\usepackage{booktabs}

\usepackage[T1]{fontenc}
\usepackage{inconsolata}

\title{Auto-Research Agent on MSA Organization to Improve Protein-Protein Complex Structure Prediction}

\author{
\textbf{Zilu Zhang}$^{1,\ast}$ \quad
\textbf{Jing Bao}$^{2,\ast}$ \quad
\textbf{Lifeng Qiao}$^{2,\ast}$ \\[0.3em]
\textbf{Sheng Xu}$^{3}$ \quad
\textbf{Qiantai Feng}$^{4}$ \quad
\textbf{Siqi Sun}$^{3,\dagger}$ \quad
\textbf{Yang Yang}$^{2,\dagger}$ \\[0.6em]
$^{1}$Beijing University of Posts and Telecommunications, Beijing, China \\
$^{2}$School of Artificial Intelligence, Shanghai Jiao Tong University, Shanghai, China \\
$^{3}$Research Institute of Intelligent Complex Systems, Fudan University, Shanghai, China \\
$^{4}$Shanghai Artificial Intelligence Laboratory, Shanghai, China \\[0.4em]
$^{\ast}$Equal contribution. \\
$^{\dagger}$Corresponding authors.
}

\iclrfinalcopy 

\begin{document}

\maketitle

\lhead{} 
\renewcommand{\headrulewidth}{0pt}

\begin{abstract}
Protein-complex structure prediction with AlphaFold 3 depends on MSA content, depth, and cross-chain pairing. 
Current pipelines apply uniform recipes, typically maximizing depth and pairing by species, yet even whether cross-chain pairing helps AF3 remains contested. 
We show that the value of both depth and pairing is strongly target-dependent, so no single recipe is optimal across interfaces. 
We introduce an autonomous agent that treats a frozen AF3 as a
black-box experimental instrument. 
For each target, the agent formulates biological hypotheses, analyzes and reorganizes the MSA, and iteratively refines sequence selection and pairing using predictor label-free signals alone.
On 1{,}015
protein--protein interfaces from three benchmarks released after AF3's
training cutoff, the agent improves overall interface DockQ over AF3's
default pipeline and fixed MSA recipes on every benchmark, while using on average
60\% fewer MSA sequences than the default. 
Gains are largest for inter-species heteromers, where species-based pairing is unreliable: mean DockQ increases from 0.465 to 0.612 on FoldBench and from 0.482 to 0.562 on HD439.
The improvements arise from both sequence construction and pairing policy, including deliberate non-pairing.
These results recast MSA design as a target-specific problem and show that an autonomous agent can adapt it at test time through biological reasoning and self-directed, confidence-guided experimentation.
\par\noindent
\textbf{Project page:} \url{https://sjtusaicaar.github.io/msa-pairing-agent/}

\end{abstract}


\section{Introduction}
\label{sec:introduction}


Deep learning has transformed protein structure prediction. AlphaFold2 (AF2)
\citep{jumper2021alphafold2} achieves near-experimental accuracy for many
proteins by exploiting co-evolutionary signals in multiple sequence alignments
(MSAs): by aligning homologous sequences, MSAs reveal conserved positions and
correlated substitutions that constrain residue--residue contacts. For
complexes, the same principle extends across chains. AlphaFold-Multimer
\citep{evans2021alphafoldmultimer} and AlphaFold3 (AF3)
\citep{abramson2024alphafold3,abramson2024alphafold3addendum} combine
unpaired MSAs for each chain with paired MSAs, in which homologs of different
chains are matched by species so that correlated substitutions across the
interface inform inter-chain geometry; RoseTTAFold \citep{baek2021rosettafold}
relies on MSA features in a similar way. 


Prediction accuracy therefore depends on MSA quality, which varies
substantially across targets: sparse, redundant, or incomplete alignments
can limit accuracy. One response is to bypass alignments altogether.
MSA-free predictors such as ESMFold \citep{lin2023esmfold} and
HelixFold-Single \citep{fang2023helixfoldsingle} instead encode
evolutionary regularities in protein language models. 
Like MSA-based methods, however, their accuracy decreases for targets with
few homologs \citep{lin2023esmfold,fang2023helixfoldsingle}.
For heteromeric complexes, recent benchmarking \citep{luo2026benchmarking}
challenges the assumption that precise cross-chain pairing is essential:
on average, increasing the depth of unpaired alignments helps AF3 as much
as or more than species-based pairing, a trend that extends to antibody--antigen
complexes. These findings, however, concern a single construction axis,
depth versus pairing, and are reported as aggregate trends; how this axis
interacts with other construction choices, and whether such trends hold
for individual targets, remains untested.

This ambiguity calls for an agent that decides, for each target, how deep to search, what to filter, and how to pair sequences, and that refines these decisions using feedback from downstream structure prediction.
Evaluator-guided LLM auto-research offers a relevant precedent. In combinatorial optimization and algorithm design, FunSearch \citep{romera-paredes2024funsearch}, EoH \citep{liu2024eoh}, ReEvo \citep{ye2024reevo}, and AlphaEvolve \citep{novikov2025alphaevolve} 
use LLMs to search over programs scored by an evaluator, discovering heuristics that outperform hand-designed ones.
These successes rest on two conditions: instances are homogeneous, so one program serves the whole problem class; and the score is the objective itself (e.g., tour length or number of bins), so optimizing it solves the task. MSA construction satisfies neither:
\begin{itemize}
\item \textbf{Target-dependent optima.} Proteins differ widely in family size and diversity of available homologs, so a strategy that works well for one target often fails to transfer to another.
\item \textbf{No ground-truth signal as guidance.} For a new target the true structure is unknown, and the only available feedback is AlphaFold3's confidence scores (e.g., pLDDT, ipTM) \citep{abramson2024alphafold3}, which are the model's own estimates of accuracy rather than measurements of it. Optimizing them directly risks exploiting their miscalibration, so they must be weighed against direct analysis of the MSA and predicted structure.
\end{itemize}
The central challenge is therefore not to find a single better heuristic, but to build a general-purpose environment in which an agent can explore, reflect, and refine a construction strategy for each new target. Our contributions are:

\begin{itemize}
\item \textbf{Auto-Research Agent.} We introduce an autonomous agentic framework that keeps the structure predictor in the loop. The agent iteratively proposes, evaluates, and refines target-specific MSAs using only the predictor's label-free feedback.

\item \textbf{SOTA Performance.} Across three protein--protein benchmarks, the agent consistently improves over AF3-Default setting and other heuristic strategies, while using substantially shallower MSAs, with average depth reduced by approximately 60\%.

\item \textbf{Target-specific MSA reasoning.} The agent's key capability is to turn target-specific biological diagnoses into an iterative hypothesis-testing process, reading the predictor's confidence feedback to refine its MSA construction step by step.

\end{itemize}

\section{Related Work}
\label{sec:related_work}
\paragraph{MSA construction and pairing.}
Building an MSA for a complex involves two decisions: how deep each chain's
alignment is, and how its rows are paired across chains.
\textit{Retrieval} tools such as JackHMMER \citep{eddy2011hmmer}, HHblits
\citep{remmert2011hhblits}, and MMseqs2 \citep{steinegger2017mmseqs2}, as used in
ColabFold \citep{mirdita2022colabfold}, determine each chain's depth and diversity.
\textit{Pairing} methods then match rows across chains, using species and genomic
proximity \citep{mirdita2022colabfold}, coupling or mutual-information objectives
\citep{bitbol2016ipa,bitbol2018miipa}, or differentiable matching
\citep{lupo2024diffpalm,lupo2025diffpass}.
Which decision matters more is contested: across hundreds of heterodimers,
most reported pairing gains stem from the accompanying increase in depth
\citep{luo2026benchmarking}.
\textit{Portfolio-based} pipelines hedge across these choices by generating
alternative inputs for each target and keeping the best: DeepMSA2
\citep{zheng2024deepmsa2} selects among MSAs from hierarchical database searches
using AlphaFold-based scores; MULTICOM \citep{liu2023multicomcasp15,liu2025multicom4complex}
builds alternative alignments, templates, and paired MSAs from multiple sources of
evidence; and CAF3 \citep{neupane2026caf3} evaluates pre-enumerated MSA and template
variants for AF3.
These pipelines adapt to each target only through selection: candidates are
produced by fixed rules, and the scores do not inform which new candidates to
generate.

\paragraph{Autonomous research agents.}
LLM agents increasingly conduct research in a propose--evaluate--revise loop.
In domains with reliable evaluators, they evolve programs and heuristics
against executable scores
\citep{romera-paredes2024funsearch,novikov2025alphaevolve,liu2024eoh,ye2024reevo}
and automate machine-learning research whose experiments return measurable
outcomes \citep{lu2024aiscientist,schmidgall2025agentlaboratory}. Because the
evaluator is trustworthy, research reduces to optimizing against it. Agents in chemistry and biology, by contrast, have primarily served a different role: orchestrating specialized domain tools to plan experiments, analyze intermediate results, or execute established research workflows
\citep{bran2024chemcrow,ghafarollahi2024protagents,shen2024toursynbio,xin2024bia,sinclair2025structbioreasoner,qu2026crisprgpt}
rather than iterate toward an objective. When they do iterate in silico, they
score candidates with computational proxies such as predicted structure
confidence \citep{swanson2025virtuallab}, implicitly treating these proxies as
ground truth; trustworthy closed-loop feedback has so far required physical
experiments \citep{boiko2023coscientist}.

\section{Preliminary}



\paragraph{Proteins, Complexes, and MSAs.}
Proteins are chains of amino acids (20 common types); writing each as a letter turns a protein into a string, its \textit{sequence}. Many proteins function as complexes of several chains, which fall into three categories: \textit{homomers} (identical chains), \textit{intra-species heteromers} (distinct chains from one organism, e.g., hemoglobin's $\alpha$ and $\beta$ chains), and \textit{inter-species heteromers} (chains from different organisms, e.g., a viral protein bound to a host receptor, or an antibody bound to an antigen). \textit{Homologs} are proteins in different species that descend from a common ancestral gene and share similar sequences and structures. A multiple sequence alignment (MSA) aligns a query's homologs so that corresponding positions share a column. Because residues in spatial contact tend to mutate together, co-varying columns indicate structural proximity. A single-chain MSA captures only intra-chain coevolution. A \textit{paired} MSA extends this signal across chains by matching each chain's homologs by source species, so each row holds proteins from one organism. Pairing therefore mainly benefits intra-species heteromers, whose chains coevolve within one organism.

\paragraph{MSA representation.}
\label{par:msa_representation}
Consider a target complex $k$ with $C$ protein chains, and let $x_c\in\Sigma^{L_c}$ be the query sequence of chain $c$, where $\Sigma$ is the amino-acid alphabet augmented with a gap token; we omit $k$ from chain-level quantities when clear. Chain $c$ has an unpaired MSA $U_c\in\Sigma^{N_c^u\times L_c}$, whose rows are homologs retrieved independently for that chain, and a paired MSA $P_c\in\Sigma^{N^p\times L_c}$. For each row index $i$, the corresponding rows across chains contain homologs associated with the same UniProt organism ID, with all-gap rows for chains lacking a corresponding homolog. These rows define a putative cross-chain correspondence. The complete MSA representation is $A=\{(U_c,P_c)\}_{c=1}^{C}$. Let $\mathcal{H}_k$ denote the homologs retrieved for all chains of target $k$. A target-specific MSA construction strategy $g_k$ pairs, orders, and combines these sequences into an MSA proposal $A_k=g_k(\mathcal{H}_k)$.


\paragraph{DockQ evaluation.}
\label{par:dockq}

We score each interface with DockQ~\citep{10.1371/journal.pone.0161879},
which lies in $[0,1]$ and combines three measures of agreement with the
reference structure: the fraction of native interface contacts recovered
($F_{\mathrm{nat}}$), the RMSD of interface residues ($\mathrm{iRMSD}$), and
the RMSD of the smaller chain after superposing the larger one
($\mathrm{LRMSD}$). 
An interface $i$ on complex target $k$ is a pair of chains in contact in
the reference structure. Given an MSA proposal $A_k$, AF3 generates $S$
structure samples with ranking scores $\{r_s\}_{s=1}^{S}$. Let $Q_{i,s}$ be
the DockQ of sample $s$ on interface $i$. We report
\[
Q_i^{\mathrm{mean}}=\frac{1}{S}\sum_{s=1}^{S} Q_{i,s},\qquad
Q_i^{\mathrm{rank}}=Q_{i,\hat s},\;\; \hat s=\arg\max_s r_s,\qquad
Q_i^{\mathrm{oracle}}=\max_s Q_{i,s},
\]
which measure average sampling quality, the quality of the structure AF3
selects, and the best structure it samples, respectively. Dataset-level
scores are macro-averaged over interfaces.

\section{Methodology}

\label{sec:methodology}

\subsection{Overview}
\label{sec:proposal-construction-loop}

We formulate the design of MSA representation as an autonomous agentic exploration process. 
The framework comprises three components: 1) the agent that iteratively proposes MSA representation; 2) the predictor $\mathcal{G}$ such as AF3 that generates structure predictions from each proposal; and 3) the runtime environment that executes tool calls, enforces budgets, and manages the prediction-derived signals observed by the agent.

Concretely, given a multichain complex target $k$, the agent constructs an initial \textbf{MSA representation proposal} $A_k^{(1)}$ that includes unpaired MSAs for every chain and paired MSAs across chains (Preliminaries~\ref{par:msa_representation}). The agent then enters an iterative refinement process. In each round $r$, it submits $A_k^{(r)}$ to $\mathcal{G}$ and receives prediction-derived feedback signals $\mathcal{F}(A_k^{(r)})$ as observables. This feedback is derived exclusively from the outputs of G and does \textit{not} require access to the ground-truth structure (Appendix~\ref{app:af3_msa_overview}). Based on this feedback, the agent may write and execute its own analysis programs before producing a refined proposal, $A_k^{(r+1)}$.

This proposal--feedback cycle continues until the submission budget is exhausted or the agent concludes further improvement is unlikely. Finally, the agent nominates one MSA representation $A_k^{*}$ from its previous proposals. We evaluate $A_k^*$ by submitting it to $\mathcal{G}$ and comparing the resulting predicted structure with the ground-truth structure. We then get the evaluation signal $\mathcal{E}(A_k^*)$ using DockQ, as defined in Preliminaries~\ref{par:dockq}.

It is worth noting that, in our main experiments, the runtime exposes only $\mathcal{F}(A_k^{(r)})$ to the agent. It computes $\mathcal{E}(A_k^{(r)})$ silently and never exposes these signals through the agent's context or workspace. As an ablation study, we consider an oracle setting in which $\mathcal{E}(A_k)$ is visible to the agent, as described in Appendix~\ref{gtgr}.

\begin{figure}[!htb]
    \centering
    \includegraphics[width=0.85\linewidth,
        height=0.38\textheight,
        keepaspectratio]{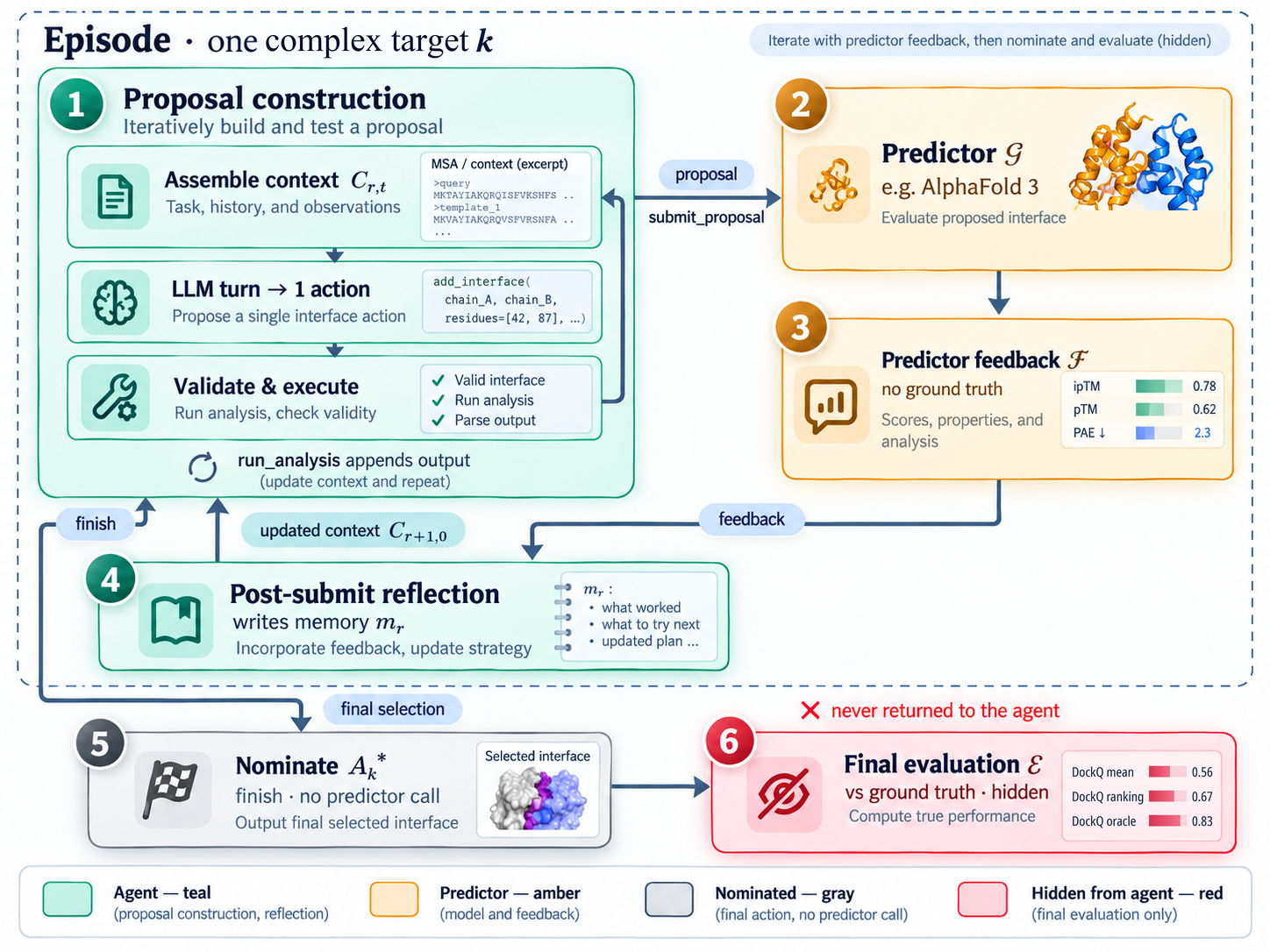}
    \caption{An autonomous agent refines per-complex MSA organization via a double loop: inner analysis builds each proposal, and outer AlphaFold 3 feedback plus reflection guides the next.}
    \label{fig:agent_framework_overview}
\end{figure}

\subsection{Autonomous Agent with Predictor-in-the-loop}
\label{sec:agent_loop}
\paragraph{Double-loop Architecture.} Each run on target $k$ operates via a hierarchical double-loop architecture across three nested levels (Fig.~\ref{fig:agent_framework_overview}); 1) An outer-loop \textbf{episode} is a sequence of proposal rounds that drives MSA representation evolution. It ends when the agent nominates one previously submitted proposal as the final proposal $A_k^{*}$; 
2) A inner-loop \textbf{proposal round} $r$ is
an optimization loop consisting of two consecutive phases. 
In \emph{Construction} phase, the agents iteratively analyzes the target and earlier findings until it successfully submit a proposal $A_k^{(r)}$ to generator $\mathcal{G}$. In \emph{Reflection} phase, the agent interprets generator feedback $\mathcal{F}(A_k^{(r)})$ and writes a structured summary $m_r$ to reflect its learning.
3) An \textbf{LLM turn} is a single model invocation, which may issue one or more actions. A proposal round typically spans many turns (Appendix~\ref{app:system_prompt}). 

Round~1 is a mandatory control that submits the default AF3 input, so the
default is always a nominable candidate and the reference for all later
proposals.

\paragraph{Actions.}
At turn $t$ of prospective round $r$, LLM is prompted to generate action $a_{r,t}$ together with the action-specific arguments $\phi_{r,t}$. The available actions depend on the phase:
\begin{equation}
    \mathcal{A}_{\text{con}} = \{\texttt{analysis}, \texttt{submit}, \texttt{finish}\},
    \qquad
    \mathcal{A}_{\text{ref}} = \{\texttt{analysis}, \texttt{submit\_round\_summary}\}.
    \label{eq:action-space}
\end{equation}
The $\mathcal{A}_{\text{con}}$ and $\mathcal{A}_{\text{ref}}$ represent the available actions in Construction and Reflection phases, respectively. The action definitions are:
\begin{itemize}
  \item \texttt{analysis} runs an agent-written Python program in the
  sandbox (Section~\ref{sec:agent-workspace}) and returns its output.
  It does not change the phase; at most $L$ calls are allowed per reflection.
  \item \texttt{submit} executes a builder script, validates the resulting
  proposal, runs $\mathcal{G}$, and returns $\mathcal{F}(A_k^{(r)})$,
  consuming one unit of budget. The round then enters reflection.
  \item \texttt{submit\_round\_summary} stores the summary $m_r$ and ends
  round $r$.
  \item \texttt{finish} nominates one submitted proposal as $A_k^{*}$ with a
  written justification and ends the episode. The agent is instructed to
  weigh proxy confidence, robustness across seeds, and biological
  plausibility. Under ground-truth feedback, the tested proposal with the
  highest mean DockQ is nominated automatically.
\end{itemize}

\paragraph{Context and Memory.}
Round $r$ runs in a single conversation spanning both Construction and Reflection phases. 
The ordered sequence of
summaries from earlier completed rounds,
$M_{<r} = (m_1, \dots, m_{r-1})$, forms the agent's
\emph{cross-round memory}; it is the \textit{only} information carried from one
round to the next. At turn $t$ of round $r$, the model context is
\begin{equation}
    C_{r,t} = \bigl(S,\ M_{<r},\ H_{r,<t}\bigr),
    \qquad
    H_{r,<t+1} = H_{r,<t} \mathbin{\Vert} \bigl(a_{r,t}, o_{r,t}\bigr),
    \label{eq:context-update}
\end{equation}
where $S$ is the fixed system prompt, $M_{<r}$ is serialized as JSON in
the first user message ($M_{<1}$ is empty), $a_{r,t}$ is the action,
$o_{r,t}$ is action's output, and $\Vert$ denotes ordered concatenation.
Target inputs are never placed in the context; the agent reads them
through \texttt{analysis}.

\subsection{Runtime and Workspace}
\label{sec:runtime}

The runtime separates what the agent sees from what it keeps: the LLM
context is reset every round (Section~\ref{sec:agent_loop}), whereas an
episode-scoped workspace persists. This section
describes the workspace, the sandbox in which analysis code runs, and how
the runtime validates actions and enforces budgets.

\paragraph{Workspace.}
\label{sec:agent-workspace}
Each episode has a private workspace. The runtime writes the target
inputs into it at the start of the episode and adds predictor feedback
after each valid submission, restricted to what the current execution
profile makes visible. The agent adds its own files, such as analysis
programs, derived statistics, notes, and candidate MSA organizations,
through \texttt{analysis}. The runtime thus supplies the target and the
predictor signal, while the agent decides how to process that signal and
what to keep. Workspace files never enter the context automatically and
must be read through \texttt{analysis}. The folder layout is given in
Appendix~\ref{workspace}

\paragraph{Analysis sandbox.}
Rather than a fixed set of analysis tools, the agent receives a
general-purpose programming environment, so it can decide which evidence
matters for a given target and compute it itself. Each \texttt{analysis} call specifies an objective and a Python
program, which runs in a fresh, isolated process. The process may read
the target inputs, visible predictor feedback, and existing workspace
files, and may write only to the agent-writable part of the current
workspace. Network access is disabled to prevent ground-truth leakage.
The process terminates after the call, and its textual output is
appended to the context as $o_{r,t}$.

\paragraph{Action validation.}
The runtime checks every action against its schema before execution,
including proposal submissions and memory entries. A failed check is
returned as the tool output and followed by another LLM turn; it changes
the context but not the proposal, the memory, or the predictor-call
count. If a turn contains no action, the runtime appends a continuation
instruction.

\paragraph{Budgets and termination.}
  Each episode has an LLM exploration budget and a submission
  budget. Every LLM turn during exploration, including reflection
  and correction turns, counts toward the former; each successfully
  completed submission counts toward the latter. Once the LLM
  exploration budget is exhausted, analysis and new submissions
  are disabled. Up to two additional LLM turns are reserved for
  any pending round summary and \texttt{finish}; these turns are
  included in the reported LLM-call total. The episode is marked
  unsuccessful if the required finalization is not completed
  within the allowed calls. 

\section{Experiment Set-up}
\subsection{Benchmark Datasets}
We evaluate on three protein--protein interface benchmarks comprising 1{,}015
interfaces in total, all released after AF3's training cutoff (details in Appendix~\ref{app:data} and \ref{app:targets}).

\textbf{FB-v1-pp} (278 interfaces; 194 homomeric, 84 heteromeric) is the protein-protein subset of the published FoldBench benchmark~\citep{xu2026foldbench}.
It excludes interfaces similar to AF3's
training data at the chain and whole-complex level, so that accuracy
reflects what AF3 infers from the input MSA rather than what it has
memorized.

\textbf{FB-v2-pp-hard} (299 interfaces; 168
homomeric, 131 heteromeric) is a hard subset of FoldBench-v2 currently under development. It retains only interfaces
whose local binding geometry and residues differ substantially from
AF3's training set.

\textbf{HD439} (438 heteromeric
interfaces) is a public benchmark ~\citep{luo2026benchmarking} comprising entirely of heterodimers: 99
inter-species and 339 intra-species pairs. 

\subsection{Agent Setup}
\label{sec:experimental_setup}
\paragraph{Predictor.}
We use AlphaFold~3 (v3.0.2) as $\mathcal{G}$ with the FoldBench
inference protocol \citep{xu2026foldbench}: 10 recycles and five
diffusion samples for each of five fixed random seeds, shared across
all targets and conditions, giving 25 structures per submission.
Within each target, all conditions also use the same query sequences
and templates, so differences in predicted structures arise from the
MSA input alone.
The default condition uses the MSAs and species-based pairing of AF3's standard data
pipeline. Agent proposals instead specify each chain's MSA row by row;
homologs placed in the same row across chains form one row of the
complex MSA, so the agent controls both which sequences are included and
which are paired across chains (Appendix~\ref{app:predictor}).

\paragraph{LLM Setting.}
The agent uses GLM-5.3 with temperature 0.2, maximum reasoning effort,
and a maximum response length of 32{,}768 tokens. Each target is
processed in an independent episode whose first proposal is the
AF3-Default control (Appendix~\ref{app:agent_config}). 

\paragraph{Budget Setting.}
Each episode allows at most 60 completed generator runs, including the
initial Default proposal, and 600 LLM calls in total. Failed or invalid submissions are not
counted, and the agent may stop early by nominating a final proposal $A_k^{*}$. The analysis Python program has a execution time limit 120s (Appendix~\ref{app:budget}). 

\subsection{MSA Input Baselines}
\label{sec:msa_input_baseline}
We compare our agent with the following baselines:
\textbf{AF3-Default} is the initial MSA proposal generated by AF3 Default pipeline. It is submitted directly to the predictor without iterative refinement or proposal selection.
\textbf{mMSA} is the fixed MSA representation strategy where only monomeric alignments are provided, without additional cross-chain pairing.
\textbf{pMSA} is the strategy in which the monomeric alignments are supplemented with cross-chain paired sequences.
\textbf{uMSA} is the strategy in which the monomeric alignments are supplemented with unpaired UniProt hits. 

\paragraph{Controlled comparisons.}
The agent and AF3-Default always start from identical MSA resources, so
Agent vs.\ AF3-Default is our primary comparison: it isolates the effect
of iterative refinement and proposal selection. The three other
baselines are controlled with respect to one another, but not always
with respect to the agent. On the FoldBench sets, they are built from
separate HHblits and UniProt searches. Because pMSA and uMSA draw on the
same pool of hits, their difference isolates the effect of pairing, but
their differences from the agent also reflect MSA source and
construction. On HD439, the agent and all baselines use the MSAs
released with the benchmark \citep{luo2026benchmarking}, whose
unpaired, paired, and UniProt-enriched variants serve as mMSA, pMSA,
and uMSA, so all comparisons there are input-matched. Implementation
details are given in Appendix~\ref{app:msa}.

\section{Results}
\subsection{Main Results}
\label{sec:main_results}
As Table~\ref{tab:main_results} shows, the agent's nominated MSA representations achieve the highest $\bar{Q}^{\text{mean}}$,
average $\bar{Q}^{\text{ranking}}$, and average $\bar{Q}^{\text{oracle}}$ across all benchmarks.
At the benchmark level, $\bar{Q}^{\text{mean}}$ rises
from 0.584 to 0.604 on FoldBench-v1-pp, from 0.556 to 0.583 on
FoldBench-v2-pp-hard, and from 0.583 to 0.624 on HD439.
In six out of the eight interface-type subsets across the three benchmarks,
and its gains on $\bar{Q}^{\text{mean}}$ over AF3-Default are statistically significant (Appendix Table~\ref{tab:selection-msa-shuffle}). 

All methods are scored on the same 25 structures from five seeds; the
agent's additional predictor runs (6.01 per target on average) are used
only for search, so the gains do not come from sampling more structures.
Because nomination uses confidence on these same structures, we also
re-predict the nominated MSAs with fresh seeds and find that the gains
persist (Appendix~\ref{seeds}).


\paragraph{Sequence construction and pairing policy.}
The agent controls both the sequences included for each chain and their pairing structure, including which sequences are intentionally left unpaired. To assess the contribution of this pairing policy, we independently shuffle the row order of each chain in the agent's proposed MSA (Appendix Table~\ref{tab:selection-msa-shuffle}). This preserves the per-chain sequence content but destroys the proposed pairing structure: selected pairs are broken, while intentionally unpaired sequences can be presented to AF3 as arbitrary pairs. The comparison therefore measures the value of the agent's complete pairing policy---both pairing and deliberate non-pairing---rather than pairing alone.

For homomers and inter-species heteromers, where informative cross-chain pairing is not expected, shuffling does not significantly change $\bar{Q}^{\text{mean}}$, indicating that the gains primarily arise from sequence construction. For intra-species heteromers, however, shuffling significantly lowers $\bar{Q}^{\text{mean}}$, showing that the agent's pairing policy provides additional benefit. On the 339 intra-species targets in HD439, preserving this policy accounts for approximately half of the gain over AF3-Default (0.014 out of 0.029), with the remainder attributable to sequence construction.

\begin{table*}[t]
\centering
\scriptsize
\setlength{\tabcolsep}{4pt}
\caption{Results across the three protein-protein benchmarks (Appendix~\ref{app:data}). The agent's final MSA nominations are compared with four baselines (Section~\ref{sec:msa_input_baseline}), using DockQ metrics (Section~\ref{par:dockq}).
  Bold and underline denote
the best and second-best reported results, respectively.
}
\label{tab:main_results}
\resizebox{\textwidth}{!}{%
\begin{tabular}{lllccccc}
\toprule
Dataset & Subset (Number of interfaces) & Metric
& Agent & AF3-Default
& \textit{mMSA} & \textit{pMSA} & \textit{uMSA} \\
\midrule
\multirow{12}{*}{FB-v1-pp}

& \multirow{3}{*}{Homomeric ($N=194$)}
& DockQ mean    & \textbf{0.554} & \underline{0.539} & 0.504 & 0.532 & 0.536 \\
&& DockQ ranking & \textbf{0.562} & \underline{0.554} & 0.511 & 0.546 & 0.546 \\
&& DockQ oracle  & \textbf{0.599} & 0.598 & 0.560 & 0.595 & \underline{0.598} \\
\cmidrule(lr){2-8}

& \multirow{3}{*}{Intra-species heteromeric ($N=65$)}
& DockQ mean    & \textbf{0.752} & \underline{0.751} & 0.720 & 0.723 & 0.721 \\
&& DockQ ranking & \underline{0.779} & \textbf{0.794} & 0.721 & 0.728 & 0.722 \\
&& DockQ oracle  & \underline{0.837} & \textbf{0.840} & 0.807 & 0.783 & 0.791 \\
\cmidrule(lr){2-8}

& \multirow{3}{*}{Inter-species heteromeric ($N=19$)}
& DockQ mean    & \textbf{0.612} & 0.465 & \underline{0.478} & 0.441 & 0.460 \\
&& DockQ ranking & \textbf{0.618} & 0.468 & \underline{0.537} & 0.441 & 0.490 \\
&& DockQ oracle  & \textbf{0.646} & 0.565 & \underline{0.595} & 0.494 & 0.553 \\
\cmidrule(lr){2-8}

& \multirow{3}{*}{Overall ($N=278$)}
& DockQ mean & \textbf{0.604} & \underline{0.584} & 0.553 & 0.570 & 0.574 \\
&& DockQ ranking& \textbf{0.617} & \underline{0.604}& 0.562 & 0.582 & 0.583   \\
&& DockQ oracle& \textbf{0.658} & \underline{0.652}& 0.620 & 0.632 & 0.640 \\

\midrule
\multirow{12}{*}{FB-v2-pp-hard}
& \multirow{3}{*}{Homomeric ($N=168$)}
& DockQ mean
& \textbf{0.599} & \underline{0.575} & 0.541 & 0.575 & 0.568 \\
&& DockQ ranking
& \textbf{0.597} & \underline{0.585} & 0.543 & 0.574 & 0.569 \\
&& DockQ oracle
& \textbf{0.658} & \underline{0.657} & 0.620 & 0.654 & 0.648 \\
\cmidrule(lr){2-8}

& \multirow{3}{*}{Intra-species heteromeric ($N=100$)}
& DockQ mean
& \textbf{0.623} & \underline{0.596} & 0.514 & 0.590 & 0.592 \\
&& DockQ ranking& \textbf{0.625} & \underline{0.602}& 0.535 & 0.601 & 0.594 \\
&& DockQ oracle& \textbf{0.680} & \underline{0.671}& 0.596 & 0.664 & 0.658 \\
\cmidrule(lr){2-8}

& \multirow{3}{*}{Inter-species heteromeric ($N=31$)}
& DockQ mean
& \textbf{0.371} & 0.323& 0.333 & 0.318 & \underline{0.342} \\
&& DockQ ranking
& \textbf{0.386} & 0.358& 0.360 & 0.329 & \underline{0.364} \\
&& DockQ oracle& 0.446 & \underline{0.474}& 0.469 & 0.400 & \textbf{0.474} \\
\cmidrule(lr){2-8}

& \multirow{3}{*}{Overall ($N=299$)}
& DockQ mean
& \textbf{0.583} & \underline{0.556}
& 0.510 & 0.553 & 0.552 \\
&& DockQ ranking
& \textbf{0.585} & \underline{0.567}
& 0.522 & 0.557 & 0.556 \\
&& DockQ oracle
& \textbf{0.643} & \underline{0.643}
& 0.596 & 0.631 & 0.633 \\

\midrule
\multirow{9}{*}{HD439}
& \multirow{3}{*}{Intra-species heteromeric ($N=339$)}
& DockQ mean    & \textbf{0.642} & 0.613 & 0.612 & \underline{0.623} & 0.614 \\
&& DockQ ranking & \textbf{0.650} & 0.628 & 0.627 & 0.630 & \underline{0.632} \\
&& DockQ oracle  & \textbf{0.688} & 0.681 & \underline{0.684} & 0.684 & 0.683 \\
\cmidrule(lr){2-8}

& \multirow{3}{*}{Inter-species heteromeric ($N=99$)}
& DockQ mean    & \textbf{0.562} & 0.482 & 0.483 & 0.496 & \underline{0.507} \\
&& DockQ ranking & \textbf{0.569} & 0.509 & 0.515 & \underline{0.542} & 0.542 \\
&& DockQ oracle  & \underline{0.613} & 0.599 & 0.596 & \textbf{0.614} & 0.605 \\
\cmidrule(lr){2-8}

& \multirow{3}{*}{Overall ($N=438$)}
& DockQ mean    & \textbf{0.624} & 0.583 & 0.583 & \underline{0.595} & 0.590 \\
&& DockQ ranking & \textbf{0.632} & 0.601 & 0.601 & 0.610 & \underline{0.612} \\
 && DockQ oracle  & \textbf{0.671} & 0.662 & 0.664 & \underline{0.668} & 0.666 \\

\bottomrule
\end{tabular}}
\end{table*}

\subsection{Construction quality matters more than depth.}
\paragraph{Depth does not explain the gains.} Among fixed heuristics, depth matters more than pairing: on the
FoldBench sets, where pMSA and uMSA share the same hits, adding them
unpaired generally outperforms pairing them, consistent with
\citet{luo2026benchmarking} on HD439. The agent reverses this intuition.
Its nominated MSAs average only 40\% of AF3-Default's depth, and half
retain fewer than 20\% of its rows (Appendix Table~\ref{tab:msa-depth}); since
AF3-Default and uMSA have similar depths, the same holds relative to
uMSA. Yet these much shallower MSAs outperform both. 
Shallowness alone is not the cause: mMSA is even shallower on FBv1 and FBv2 
but yields lower DockQ (Table~\ref{tab:main_results}). What matters is which 
rows are kept, not how many.

\begin{wraptable}{r}{0.58\textwidth}
  \centering
  \caption{MSA depth across benchmarks.}
  \label{tab:msa-depth}

  \resizebox{\linewidth}{!}{%
  \begin{tabular}{lrrrrrr}
    \toprule
    & \multicolumn{2}{c}{FBv1}
    & \multicolumn{2}{c}{FBv2}
    & \multicolumn{2}{c}{HD439} \\
    \cmidrule(lr){2-3}
    \cmidrule(lr){4-5}
    \cmidrule(lr){6-7}
    Method & Mean & Median & Mean & Median & Mean & Median \\
    \midrule
    AF3-Default & 5,744.5 & 2,244.5 & 10,928.3 & 14,254.0 & 8,940.7 & 9,155.5 \\
    Agent       & 2,248.4 & 428.5   & 4,097.9  & 1,025.0  & 3,417.9 & 971.5 \\
    pMSA        & 3,363.0 & 1,767.5 & 5,447.2  & 4,429.0  & 9,199.6 & 9,423.5 \\
    uMSA        & 5,213.1 & 2,056.5 & 10,296.9 & 12,334.0 & 11,613.9 & 16,384.0 \\
    mMSA        & 904.6   & 1,073.0 & 1,358.6  & 1,357.0  & 7,474.5 & 7,217.5 \\
    \bottomrule
  \end{tabular}%
  }
\end{wraptable}

\paragraph{From biological judgment to target-specific MSA proposals.}
Analyzing agent's episode trajectories revealed a common pattern:  the agent grounds its decisions in biological knowledge and, through iterative submissions, distills them into explicit and reproducible selection procedures.

For example, 8P9C is a complex comprising the tetramerization domains of p63 and p73, two paralogous members of the p53 family, and DARPin 1810 F11, a designed ankyrin repeat protein with no natural orthologs~\cite{Strubel2023DARPins}. Its homologs therefore come from the large natural ankyrin-repeat family. In the default input, 84.5\% of the 16,384 rows contained only DARPin sequences, resulting in a mean DockQ of 0.048. The agent separated p63 and p73 candidates to limit paralog mixing, paired them by species, and restricted DARPin sequences to a small, filtered supplement. The selected MSA contained only 425 rows and reached a mean DockQ of 0.866.

Another example, 7R86, pairs the leucine-rich repeat domain of the mouse Nogo receptor RTN4R with the third thrombospondin type 1 repeat of the adhesion GPCR BAI1~\cite{Wang2021RTN4}. Both are widespread repeat modules, so a single species can contribute many candidates for each chain. Species matching therefore yielded 8,191 pairs, while chain B lacked an unpaired alignment (mean DockQ 0.037). The agent kept only 78 high-identity pairs as likely orthologs and recovered 472 B-only rows from the pairing candidate pool, reaching a mean DockQ 0.921.

Across examples, the agent applied a shared biological reasoning framework but adapted its MSA construction to each target. The agent can, for example in the case of 8P9C, design its own ablations to show that both expanding the DARPin supplement and removing selected pairings substantially degraded performance. Notably, agent resolved these choices using predictor confidence alone, without access to DockQ. This demonstrates the agent's ability to turn biological hypotheses into target-specific decisions through iterative experimentation.

\subsection{Exploration Gain and Nomination Loss.}
The agent's final result depends on two abilities: finding a better MSA
during exploration, and picking a good one at the end.
For each interface, let $D$ be the mean DockQ of AF3 Default, $B$ the
highest mean DockQ among all proposals the agent tested, and $N$ the mean
DockQ of the proposal it nominates.
We call $B - D$ the \textbf{exploration gain}, the largest improvement the
agent found, and $B - N$ the \textbf{nomination regret}, the part of that
improvement lost when choosing.
The final improvement is their difference:
\begin{equation}
  N - D = (B - D) - (B - N).
\end{equation}

\begin{figure}[!htb]
      \centering
    \includegraphics[width=\linewidth]{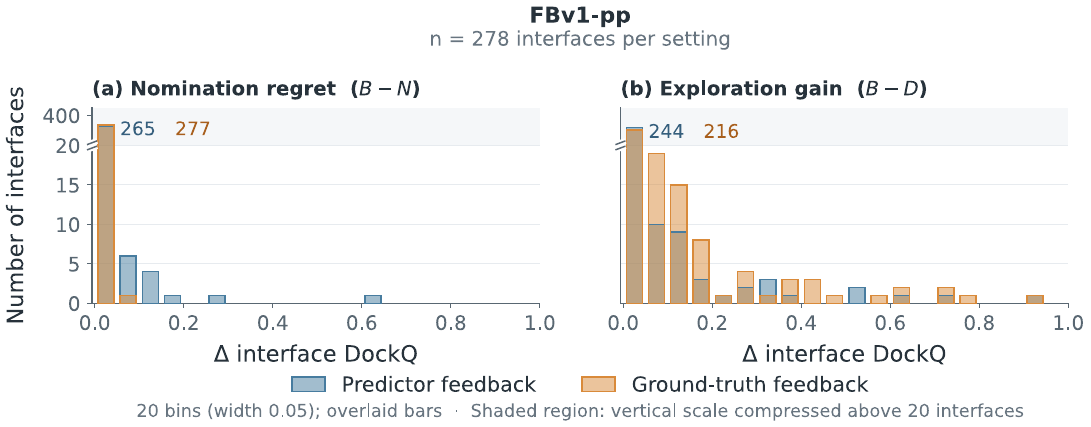}
 \caption{
  \textbf{Exploration and nomination on FBv1-pp.}
  Semi-transparent histograms compare the predictor-feedback and
  ground-truth-guided settings on the same 238 targets:
  (a) nomination regret ($B-N$) and
  (b) exploration gain ($B-D$). More
   predictor-feedback results on FBv2-pp-hard or HD439 are shown in
  Appendix Figure~\ref{fig:exploration-selection_2hd}.
  }
      \label{fig:exploration-selection}
  \end{figure}

Figure~\ref{fig:exploration-selection} compares both quantities on 278
interfaces from 238 matched FBv1-pp targets under two settings: the main
setting with \textbf{predictor feedback}, and an oracle setting with
\textbf{ground-truth feedback}, in which the agent also sees DockQ during
the search and its best tested proposal is nominated automatically.

With predictor feedback, both steps limit the final result. Using DockQ $Q^{\text{mean}}$
as metric, the exploration gain exceeds 0.05 on 34 of the 278 interfaces,
and the nomination regret exceeds 0.05 on 13 interfaces.
On average, the best tested proposal is 0.035 above AF3 Default,
but nomination gives back 41.0\% of this gain, leaving a final improvement of 0.020.

With ground-truth feedback, the agent always nominates its best proposal (target-level),
so the interface-level nomination regret is almost zero and the exploration gain equals the
final gain.
The exploration gain itself also rises from 0.035 to 0.059 (71.7\%).
A more accurate feedback signal thus improves both steps: it helps the
agent search more effectively and removes the loss at nomination.
It also suggests that the fidelity of predictor feedback is a key factor limiting agent performance, and
improving it is a promising direction for strengthening both exploration and nomination.

\FloatBarrier

\section{Conclusion}
We introduced an autonomous research agent that optimizes MSA representation for protein-complex structure prediction at test time. Using a frozen predictor’s confidence signals, the agent writes and executes analyses, accumulates evidence in memory, and adapts unpaired MSA depth and cross-chain pairing for each interface without ground-truth structures or gradients. Across FoldBench-v1, FoldBench-v2, and HD439, this approach improves overall average accuracy and confidence-based model selection, with smaller gains in best-case accuracy. The discovered strategies also provide insight into the target-dependent value of pairing and depth. Together, these findings establish adaptive MSA representation as an effective training-free approach and demonstrate how autonomous agents can use model feedback to investigate and improve inference in structural biology.

\bibliography{iclr2027_conference}
\bibliographystyle{iclr2027_conference}
\clearpage
\appendix
\section{How Alphafold 3 consumes MSA signals and provide feedback signal to the agent.}
\label{app:af3_msa_overview}
\begin{figure}[H]
    \centering
    \includegraphics[width=\linewidth]{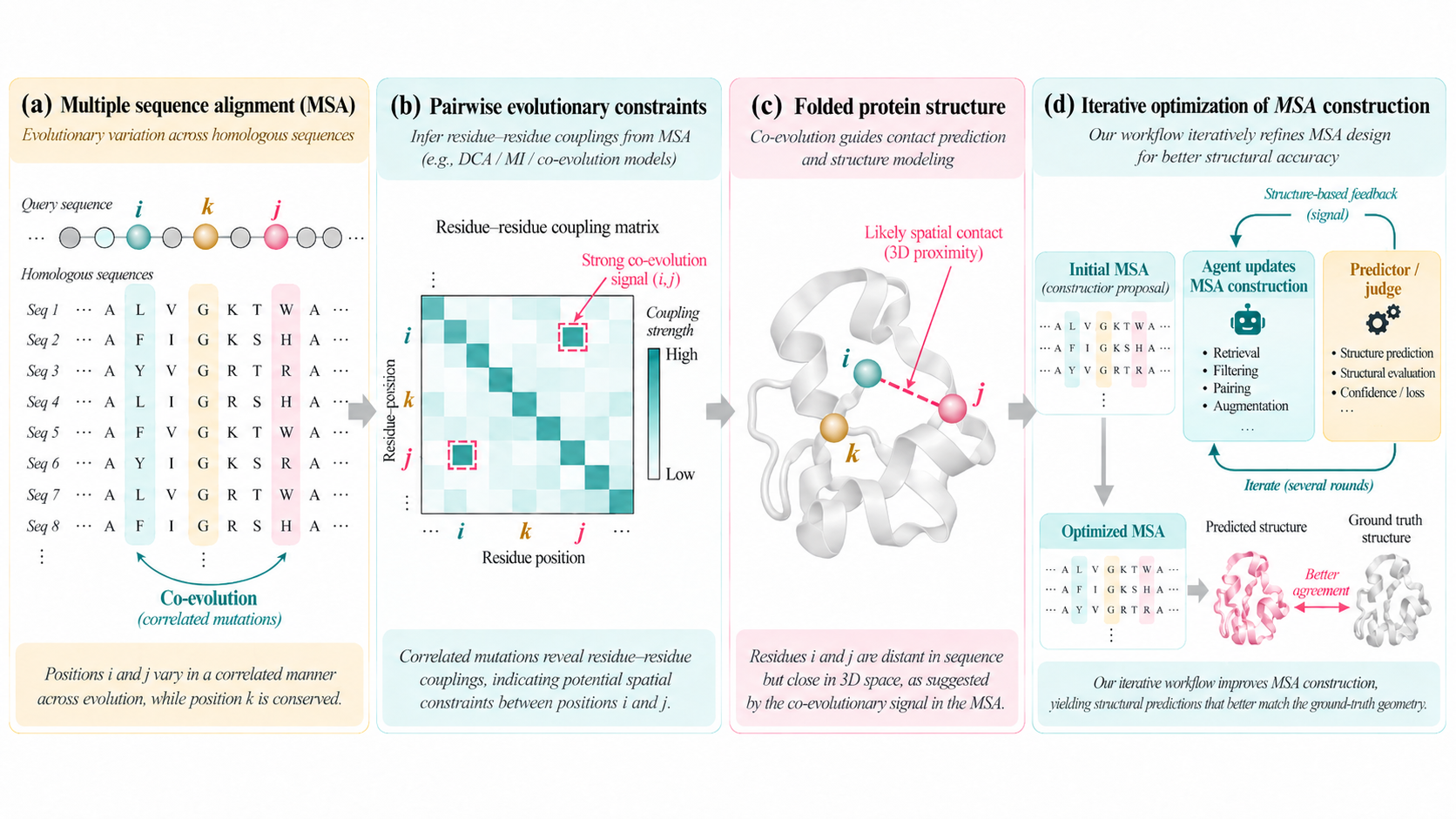}
    \caption{
        Overview of MSA-derived evolutionary signals and the
        iterative MSA construction workflow.
    }
    \label{fig:msa_overview}
\end{figure}

Figure~\ref{fig:msa_overview} provides an overview of how MSA-derived
evolutionary signals are consumed by AlphaFold~3 and how the resulting
predictor-native feedback drives our iterative MSA construction workflow.

\paragraph{AF3 feedback signal.}
We use AlphaFold~3 (AF3)~\citep{abramson2024alphafold3,abramson2024alphafold3addendum}
as a frozen structure predictor.
For each target, $X$ denotes the query sequences and $T$ the supplied
per-chain structural templates and their mappings to the query sequences.
Templates are taken from the precomputed AF3 inputs and held fixed
throughout the episode, with $T = \varnothing$ when no templates are supplied.
Given $X$, $T$, and a candidate MSA $M_g$, AF3 generates $S$ structures
$\{\hat{Y}_s\}_{s=1}^{S}$ across model seeds and diffusion samples,
together with its native confidence outputs.
Structured feedback reports the ensemble mean and maximum of ipTM, pTM,
and the native AF3 ranking score.
The agent can additionally inspect per-sample atomic coordinates,
per-chain and chain-pair confidence scores, predicted aligned error (PAE),
per-atom pLDDT, contact probabilities, disorder fractions, and clash indicators.
Rather than optimizing a predefined scalar surrogate, the agent reasons over
these predictor-native signals as proxy evidence for structural accuracy,
considering interface confidence, consistency across samples, biological and
structural plausibility, and counterevidence when proposing subsequent MSA
configurations and selecting among previously evaluated candidates.
We impose no fixed scalar combination of these signals and no rule requiring
maximization of mean ipTM.
In the blind setting, neither search nor final selection has access to
ground-truth target structures or evaluation metrics such as DockQ;
these are used only for offline evaluation.
\paragraph{MSA organization episode.}
Algorithm~\ref{alg:episode} summarizes how the agent uses this feedback within a target-specific episode. In each round, the agent inspects the current context and persistent memory, analyzes available evidence, and constructs an MSA candidate for AF3 evaluation. Validation and prediction failures are returned to the agent for revision. After a successful evaluation, the agent examines the AF3 feedback, records its interpretation in round memory, and proceeds to the next round. When the agent finishes, it selects a previously evaluated candidate; if the evaluation budget is exhausted without a valid final selection, the episode returns failure.

\begin{algorithm}[t]
\caption{MSA organization episode for target $k$}
\label{alg:episode}
\begin{algorithmic}[1]
\Require agent $\pi$; predictor $\mathcal{G}$; evaluation budget $B$
\State $\mathcal{M} \gets [\,]$ \Comment{persistent round memory}
\State $\mathcal{C} \gets$ initial target context
\State $r \gets 0$

\While{$r < B$}

    \Statex
    \State \textbf{// Phase 1: Analyze and construct a candidate}
    \Repeat
        \State $(a,\theta) \gets \pi(\mathcal{C}, \mathcal{M})$

        \If{$a=\texttt{analysis}$}
            \State $o \gets \textsc{Exec}(\theta)$
            \State $\mathcal{C} \gets \mathcal{C} \oplus (a,\theta,o)$

        \ElsIf{$a=\texttt{finish}$}
            \State \Return previously submitted candidate selected by $\theta$

        \ElsIf{$a=\texttt{submit}$}
            \State $A \gets \textsc{Build}(\theta)$
        \EndIf

    \Until{$a=\texttt{submit}$}

    \Statex
    \State \textbf{// Phase 2: Validate and evaluate the candidate}
    \If{$A$ is invalid}
        \State $\mathcal{C} \gets \mathcal{C} \oplus \text{validation error}$
        \State \textbf{continue}
    \EndIf

    \State $Z \gets \mathcal{G}(A)$
    \If{prediction fails}
        \State $\mathcal{C} \gets \mathcal{C} \oplus \text{runtime error}$
        \State \textbf{continue}
    \EndIf

    \State $r \gets r+1$
    \State $f_r \gets \mathcal{F}(Z)$
        \Comment{profile-visible AF3 feedback}
    \State $\mathcal{C} \gets \mathcal{C} \oplus (A,f_r)$

    \Statex
    \State \textbf{// Phase 3: Reflect on the completed round}
    \Repeat
        \State $(a,\theta) \gets \pi(\mathcal{C},\mathcal{M})$
        \If{$a=\texttt{analysis}$}
            \State $o \gets \textsc{Exec}(\theta)$
            \State $\mathcal{C} \gets \mathcal{C} \oplus (a,\theta,o)$
        \Else
            \State $m_r \gets \textsc{ParseMemory}(\theta)$
        \EndIf
    \Until{$m_r$ is valid}

    \Statex
    \State \textbf{// Phase 4: Commit memory and reset context}
    \State $\mathcal{M} \gets \mathcal{M} \oplus m_r$
    \State $\mathcal{C} \gets \textsc{Reset}(\mathcal{M})$

\EndWhile

\State \Return failure \Comment{budget exhausted without valid finish}
\end{algorithmic}
\end{algorithm}

\section{Experimental Details}
\label{app:exp_details}

\subsection{Prediction Targets and Interface Definition}
\label{app:targets}

A prediction target $k$ is the complete complex in one AF3 input. The agent optimizes MSA organization for the complex, so all evaluated interfaces within that target share one trajectory and final selection $A_k^*$. Each benchmark-listed interface is a separate evaluation entry, identified by its reference assembly and a pair of chain-copy identifiers. For evaluation, predicted chain copies are matched to their corresponding copies in the reference structure. Copies of the same protein entry share the constructed MSA but retain distinct chain identifiers for evaluation. The agent sees the complete complex, without benchmark interface annotations or knowledge of which interfaces are evaluated.

\subsection{Benchmark Datasets}
\label{app:data}

Table~\ref{tab:app_datasets} summarizes the evaluated subsets.
All results are computed at the interface level and reported
separately for each dataset.

\begin{table}[h]
\centering
\caption{Composition of the evaluated benchmark subsets.}
\label{tab:app_datasets}
\begin{tabular}{lcccccc}
\toprule
Dataset & Targets & Interfaces & Homo. & Intra-sp. hetero. & Inter-sp. hetero. \\
\midrule
FoldBench-v1 & 238  & 278 & 194  & 65  & 19 \\
FoldBench-v2 (hard) & 283 & 299 & 168 & 100 & 31 \\
HD439 & 438 & 438 & 0 & 339 & 99 \\
\bottomrule
\end{tabular}
\end{table}

\paragraph{FoldBench-v1.}
FoldBench-v1 is the protein--protein subset of FoldBench
\citep{xu2026foldbench}. To limit overlap with AF3's training data,
FoldBench includes only PDB entries deposited between 2023-01-13 (AF3's
validation cutoff, later than its 2021-09-30 training cutoff) and
2024-11-01, and removes interfaces similar in sequence or structure to a
reconstructed AF3 training set. The subset contains 279 interfaces from
239 targets. We exclude 8bvk, whose single homomeric interface could not
be processed because LLM API content filtering interrupted the agent,
from all methods. The evaluation set thus contains 278 interfaces from
238 targets: 194 homomeric and 84 heteromeric (65 intra-species, 19
inter-species).

\paragraph{FoldBench-v2.}
FoldBench-v2 is an extension of FoldBench currently under development.
For each interface released after the AF3 training cutoff, its
similarity to eligible pre-cutoff reference structures is estimated
using interface sequence identity and structural mapping coverage,
rather than global sequence similarity. For a candidate reference,
the score is the mean of the identity--coverage products on the two
sides of the interface; the interface similarity $S$ is the maximum
score over eligible direct and swapped chain matches. This is a proxy
based on retrieved pre-cutoff references, not an exact measure of
similarity to AF3's training set.

The benchmark samples 100 interfaces from each 0.1-wide similarity
bin over $[0,1]$. We evaluate its hard subset, defined by the strict
criterion $S<0.3$. One interface at the boundary ($S=0.300$) is
excluded, leaving 299 interfaces from 283 assemblies: 168 homomeric,
100 intra-species heteromeric, and 31 inter-species heteromeric
interfaces.





\paragraph{HD439.}
HD439~\citep{luo2026benchmarking} contains 439 heterodimeric targets, each contributing one interface: 340 intra-species and 99 inter-species. We exclude \texttt{7acw}, an intra-species target whose official MSA is unavailable, leaving 438 targets and 438 interfaces (339 intra-species and 99 inter-species). Based on PDB initial release records, the structures were released between 2021-10-06 and 2024-08-14, all after AF3's training cutoff of 2021-09-30.

\paragraph{Dataset overlaps.} 
At the PDB-entry level, our evaluated FoldBench-v1 and FoldBench-v2 sets share 15 entries, FoldBench-v1 and HD439 share 17 entries, and FoldBench-v2 and HD439 share 10 entries. Three entries are present in all three benchmarks.
We retain overlapping entries in their respective benchmarks without special treatment. For each benchmark experiment, we independently run the agent on every target, including those shared with other benchmarks, and report results separately for each benchmark.


\subsection{MSA Input Baselines}
\label{app:msa}

\paragraph{Default.}
Paired and unpaired MSAs are generated using the AlphaFold~3 v3.0.2 data pipeline with UniRef90 (\texttt{2022\_05}), MGnify (\texttt{2022\_05}), UniProt (\texttt{2021\_04}), and Small BFD (\texttt{bfd-first\_non\_consensus\_sequences.fasta}), and processed by AF3's built-in pairing procedure (Appendix~\ref{app:predictor}). This condition also serves as the agent's initial proposal $A_k^{(1)}$.

\paragraph{Controlled baselines on FoldBench.}
The controlled baselines share a monomeric alignment generated by HHblits against Uniclust30 (\texttt{2018\_08}) with an E-value threshold of $10^{-3}$, minimum query coverage of 50\%, maximum sequence identity of 90\%, and two iterations. Additional hits are retrieved from UniProt (\texttt{2021\_04}) using one JackHMMER iteration with reporting and inclusion E-value thresholds of $10^{-4}$ \texttt{max\_sequences=50000}. Both pMSA and uMSA use the same merged pool of raw search hits, without an additional global hit cap during baseline construction.
\begin{itemize}
\item \textit{mMSA}: the monomeric alignment only.
\item \textit{pMSA}: the monomeric alignment supplemented with complete cross-chain pairs. Hits are grouped by exact header TaxID without taxonomic-rank normalization, retaining only TaxIDs represented in every protein entity. Within each group, hits are matched by their order in the merged alignment: the $j$th hit of each entity forms one paired row, up to the smallest group size across entities. Candidate rows are sorted by the product of their original per-entity row indices, with ties broken by the index sum, within-group rank, and TaxID. The first 8,191 non-query paired rows are retained, or all rows if fewer are available.
\item \textit{uMSA}: the monomeric alignment combined with the same UniProt hit pool used for pMSA, without pairing, and filtered using HHfilter at 99\% sequence identity.
\end{itemize}


\paragraph{Controlled baselines on HD439.}
We use the MSA files publicly released with the HD439 benchmark as
fixed baseline inputs and do not regenerate them. The released
conditions correspond to the authors' monomer/unpaired, species-paired,
and UniProt-enriched constructions, which we denote as mMSA, pMSA, and
uMSA, respectively. In the released files, these conditions are stored
under the \texttt{unpaired}, \texttt{paired}, and \texttt{uniprot}
variants. According to the original paper, the underlying searches use
HHblits against Uniclust30 (2018\_08) and MMseqs2 against UniRef30
(2022\_02) and the ColabFold environmental database (2021\_08).
The paper reports HHblits candidate settings spanning two iterations,
E-values of $10^{-40}$, $10^{-10}$, $10^{-3}$, and $1$, coverage
thresholds of $50\%$ and $75\%$, and identity thresholds of $90\%$ and
$100\%$, with up to 12 selected candidates. It does not provide complete
command lines or all details of the AlphaFold2-based candidate-selection
procedure; therefore, we treat the released MSAs as the authoritative
baseline inputs rather than attempting an independent reconstruction.

\paragraph{Source mismatch between agent and controlled baselines.}
 On HD439, the agent, Default, and fixed MSA baselines all use
  the official precomputed MSA resources.
  The agent and Default use initial inputs constructed from
  \texttt{comb\_filter.a3m}, with rows assigned to the paired
  or unpaired channel according to whether AF3 can parse a
  species identifier.
  The uMSA baseline places the same combined alignment entirely
  in the unpaired channel; mMSA uses the supplied monomer
  alignments, and pMSA additionally uses the supplied paired
  alignments.
  On FBv1 and FBv2, the agent and Default share the same
  precomputed input MSAs, whereas the fixed baselines use
  separately constructed Uniclust30 and UniProt-based MSAs.
  The Agent--Default comparison therefore evaluates the gains
  from iterative MSA editing and submission selection with
  matched initial MSA resources.
  Comparisons with the fixed baselines on FoldBench additionally
  reflect differences in MSA source and construction, so their
  performance differences cannot be attributed solely to the
  agent.

\subsection{Predictor Configuration}
\label{app:predictor}
We use AlphaFold~3 v3.0.2 with 10 recycles and five diffusion
samples for each of five random seeds
(\texttt{42}, \texttt{66}, \texttt{101}, \texttt{2024}, and
\texttt{8888}), yielding 25 structures per complex and MSA
condition on one NVIDIA RTX 5880 Ada GPU (48 GB). A 25-structure
prediction took 8.0 minutes on average, including initialization,
compilation when required, inference, and output writing, but
excluding MSA search, agent execution, and structural evaluation.
Query sequences, seeds, inference settings, and template inputs
are fixed across conditions for each complex. All predictions
use precomputed inputs with the AF3 data pipeline disabled
during inference. The aggregate AF3 inference GPU hours for each dataset are reported in
Table~\ref{tab:agent_resource_usage}. 
 \begin{table}[H]
  \centering
  \caption{Resource usage of the agent episodes included in the main
  comparison. Submission counts include the initial Default. LLM calls
  count logged provider invocations, including failed and interrupted
  calls; retries within an invocation are not counted separately.
  Token totals include API-reported input and output tokens, including
  cached inputs. GPU hours cover AF3 model inference for successful
  submissions only.}
  \label{tab:agent_resource_usage}
  \small
  \setlength{\tabcolsep}{5pt}
  \begin{tabular}{lrrrrrrr}
  \toprule
  Dataset & Targets
  & \multicolumn{2}{c}{Submissions per target}
  & \multicolumn{2}{c}{LLM calls per target}
  & \shortstack{Total tokens\\($10^6$)}
  & \shortstack{AF3 inference\\(GPU hours)} \\
  \cmidrule(lr){3-4}\cmidrule(lr){5-6}
  & & Mean & Median & Mean & Median & & \\
  \midrule
  FoldBench-v1 & 238 & 6.55 & 6 & 115.11 & 109 & 840.35   & 156.07 \\
  FoldBench-v2 & 283 & 7.31 & 7 & 134.50 & 126 & 1,248.56 & 320.91 \\
  HD439       & 438 & 7.06 & 7 & 140.86 & 134 & 1,956.43 & 348.87 \\
  \bottomrule
  \end{tabular}
  \end{table}

AF3-Default retains the pipeline-generated \texttt{pairedMsa}
and \texttt{unpairedMsa} fields and AF3's standard featurization.
For the agent and
the controlled baselines (\textit{mMSA}, \textit{pMSA}, and
\textit{uMSA}), we instead specify the cross-chain row layout
explicitly: matched homologs occupy corresponding rows, while
all-gap rows represent chain-specific homologs. These layouts
are supplied through \texttt{unpairedMsa}, with
\texttt{pairedMsa=""} and
\texttt{resolve\_msa\_overlaps=false}. The initial
\texttt{backend\_default} proposal preserves both input MSA
channels; custom \texttt{manual\_row\_aligned} proposals may
draw from either channel but write their final layout only to
\texttt{unpairedMsa}.

For FBv1 and FBv2, templates were obtained by the AF3 v3.0.2
data pipeline using HMMsearch against the 2022-09-28 PDB SEQRES
snapshot and PDB mmCIF structures, with a maximum template
release date of 2023-01-12. Template structures and
query-to-template mappings are held fixed across conditions.
For HD439, no structural templates are supplied.

\subsection{Agent Configuration}
\label{app:agent_config}

\paragraph{Language model.}
We use GLM-5.3 with temperature 0.2, reasoning effort set to
\texttt{max}, and a maximum response length of 32,768 tokens.
Each target is processed in an independent episode with no
information shared across targets.
  We accessed the model via the Zhipu GLM Coding API during
  September 11--22, 2026 (UTC+8), with the API-reported model
  identifier \texttt{glm-5.3}. Table~\ref{tab:agent_resource_usage} reports the mean and median
numbers of LLM calls per target, as well as the total token usage
for each dataset.

\paragraph{Initial proposal.}
The initial proposal $A_k^{(0)}$ is the Default condition, i.e.,
AF3’s default processing of the prepared source MSAs for FoldBench-v1 and FoldBench-v2, and the publicly released MSA files for HD439. Query
sequences, templates, and predictor settings remain fixed across
all subsequent proposals.


\paragraph{Proposal generation.}
The LLM writes complete Python programs to select, filter, combine, reorder, or pair MSA rows using available inputs and accumulated observations. No fixed transformation library or scalar reward is imposed; the agent uses AF3 confidence, cross-sample consistency, and structural and biological plausibility to guide its search.

Before AF3 inference, the host validates the candidate JSON and required protein entities. For manually constructed MSAs, it checks query headers, alignment lengths, equal row counts across chains, and the specified paired-row correspondence. Gap-only rows are permitted for unpaired homologs. Invalid candidates are returned to the agent for correction without running AF3.

Programs run in an isolated Python sandbox with a 120-second limit per execution. The agent can access provided AF3 inputs, earlier observations, and prediction artifacts, but has no network access or access to ground-truth structures and evaluation results.

\paragraph{Final nomination.}
At termination, the LLM nominates one successfully evaluated
proposal $A_k^*$, with the initial Default as an eligible choice,
and provides a rationale. Conflicts among proxy metrics are
resolved by the LLM's contextual judgment, without prescribed
metric weights or priorities. The prediction ensemble already
computed for the nominated proposal is used for final evaluation;
no additional inference is run. The mean and median numbers of successful submissions per target,
including the initial Default, are reported in
Table~\ref{tab:agent_resource_usage}.

\paragraph{Isolation from ground truth.}
Ground-truth structures and all reference-based evaluation results,
including DockQ, are withheld from the agent throughout exploration
and nomination. DockQ is computed separately and used exclusively
for offline evaluation.

\subsection{Budgets and Accounting}
\label{app:budget}
Each episode permits at most 60 successful predictor submissions,
including the initial Default, and 600 agent-side LLM calls during
exploration. A submission is counted only after input validation and
successful predictor inference; invalid inputs and failed predictions
do not count, while an invocation that succeeds after internal retries
counts once. All exploration-stage LLM calls count toward the
600-call limit, regardless of purpose. Backend or service retries
within a single provider invocation do not count as separate calls.

If this LLM limit is reached before a valid final nomination, analysis
and predictor submissions stop. Up to two additional LLM calls are
allowed solely to finalize the episode: the agent first completes any
pending round summary, then nominates a previously completed proposal
via \texttt{finish}. If no summary is pending, it nominates directly
and may retry once if the nomination is invalid. The episode ends
upon a valid nomination.



\subsection{Evaluation Metrics}
\label{app:metrics}
  Interface accuracy is measured with DockQ, computed using the
  DockQ implementation in OpenStructure~2.11.1 through the
  FoldBench evaluator and its reference-to-prediction chain
  mapping (Appendix~\ref{app:targets}). For each interface and experimental
  condition, the primary reported DockQ value is the arithmetic
  mean of the scores across all 25 predicted structures
  (five seeds with five samples per seed). The same aggregation
  is applied to all experimental conditions.


\section{Supplementary Results}
 For the comparisons in Table~\ref{tab:selection-msa-shuffle},
 we compare matched per-interface mean DockQ values
within each dataset and interface type using two-sided paired
$t$-tests (Agent minus Shuffled and Agent minus AF3-Default).
Interfaces, including those from the same target, are treated as
independent; We report two-sided $p$-values; each test uses $N-1$ degrees of
freedom, where $N$ is the number of matched interfaces in that group.

Agent has higher mean DockQ than AF3-Default in all eight groups,
although the gains are small for FBv1 homomers and intra-species
heteromers. Differences from Shuffled are less consistent, including
a slight decrease for FBv2 inter-species heteromers. These results
indicate that the gains over AF3-Default cannot be uniformly
attributed to preserved cross-chain pairing.
  
Table~\ref{tab:shallow_msa_interface_gains} shows 25 interfaces for which the Agent selected a substantially shallower MSA ($N_{\mathrm{Agent}}\leq512$ versus $N_{\mathrm{Default}}>1{,}000$) and improved mean DockQ by at least 0.10. These cases span all three benchmarks and several interface types, illustrating that greater MSA depth does not always yield better structure predictions.

Table~\ref{tab:exploration-selection-cases-all} illustrates both the potential and the limitation of the selection process. In some examples, the Agent identifies an improved submission and selects it; in others, an improvement is found but missed at final selection. The Agent can also retain the Default when no tested alternative improves it, although some selections regress. These examples are illustrative and do not represent outcome frequencies across the benchmarks.

\begin{table*}[t]
\centering
\small
\setlength{\tabcolsep}{3pt}
\caption{Mean interface DockQ for the Agent-selected submission
(Agent), its shuffled-MSA counterpart (Shuffled), and the initial
native-control submission (AF3-Default).
Each per-interface score averages DockQ over 25 predicted structures;
group means weight the $N$ interfaces equally.
$\Delta\bar{Q}^{\mathrm{mean}}_{\mathrm{AS}}$ and
$\Delta\bar{Q}^{\mathrm{mean}}_{\mathrm{AA}}$ denote Agent minus
Shuffled and Agent minus AF3-Default, respectively.
Parentheses in the difference columns report two-sided $p$-values
from paired $t$-tests on matched per-interface mean scores,
treating interfaces as independent with $N-1$ degrees of freedom.
The $p$-values are unadjusted for multiple comparisons.
One MSA-shuffling realization is used per target; protein inputs
identical apart from chain IDs share the same permutation.}
\label{tab:selection-msa-shuffle}
\begin{tabular}{llrrrrr}
\toprule
Dataset & Interface type & Agent & Shuffled & AF3-Default &$\Delta\bar{Q}^{\mathrm{mean}}_{\mathrm{AS}}$ ($p$) & $\Delta\bar{Q}^{\mathrm{mean}}_{\mathrm{AA}}$ ($p$) \\
\midrule
\multirow{3}{*}{FBv1-pp} & Homomeric (N=194) & 0.554 & 0.552 & 0.539 & $+0.002\;(0.32)$ & $+0.014\;(0.00)$ \\
 & Intra-species heteromeric (N=65) & 0.752 & 0.745 & 0.751 & $+0.007\;(0.06)$ & $+0.001\;(0.87)$ \\
 & Inter-species heteromeric (N=19) & 0.612 & 0.526 & 0.465 & $+0.086\;(0.17)$ & $+0.147\;(0.03)$ \\
\midrule
\multirow{3}{*}{FBv2-pp-hard} & Homomeric (N=168) & 0.599 & 0.597 & 0.575 & $+0.002\;(0.42)$ & $+0.024\;(0.00)$ \\
 & Intra-species heteromeric (N=100) & 0.623 & 0.598 & 0.596 & $+0.025\;(0.05)$ & $+0.027\;(0.01)$ \\
 & Inter-species heteromeric (N=31) & 0.371 & 0.380 & 0.322 & $-0.010\;(0.44)$ & $+0.048\;(0.12)$ \\
\midrule
\multirow{2}{*}{HD439} & Intra-species heteromeric (N=339) & 0.642 & 0.628 & 0.613 & $+0.014\;(0.00)$ & $+0.029\;(0.00)$ \\
 & Inter-species heteromeric (N=99) & 0.562 & 0.537 & 0.482 & $+0.025\;(0.03)$ & $+0.079\;(0.00)$ \\
\bottomrule
\end{tabular}
\end{table*}

\begin{table*}[t]
\centering
\scriptsize
\setlength{\tabcolsep}{4pt}
\caption{Interfaces with shallow Agent-selected MSAs and improved DockQ.
All evaluated interfaces satisfying $N_{\mathrm{Default}}>1{,}000$,
$N_{\mathrm{Agent}}\leq512$, and $\Delta\mathrm{DockQ}\geq0.10$ are shown.
MSA depth is the number of AF3 MSA feature rows after pairing, configured
overlap handling, merging, and cropping, but before padding; query rows are included.
DockQ is averaged over 25 structures (five seeds and five diffusion samples per seed)
for each interface using the corresponding final Agent selection or AF3-Default.
$\Delta$ is Agent minus AF3-Default, computed before rounding.
Interfaces are identified by target and chain IDs; FoldBench entries use assembly1.
Interfaces from the same target share MSA depths.}
\label{tab:shallow_msa_interface_gains}
\resizebox{\textwidth}{!}{%
\begin{tabular}{lllrrrrr}
\toprule
\multirow{2}{*}{Dataset} & \multirow{2}{*}{Interface}
& \multirow{2}{*}{Interface type}
& \multicolumn{2}{c}{MSA depth}
& \multicolumn{2}{c}{DockQ mean}
& \multirow{2}{*}{$\Delta$} \\
\cmidrule(lr){4-5}\cmidrule(lr){6-7}
&&& AF3-Default & Agent & AF3-Default & Agent & \\
\midrule
\multirow{9}{*}{FBv1} & \texttt{8P9C} (\texttt{B}--\texttt{C}) & Inter-species heteromeric & 16,384 & 425 & 0.02936 & 0.94552 & +0.91616 \\
 & \texttt{8P9C} (\texttt{A}--\texttt{C}) & Inter-species heteromeric & 16,384 & 425 & 0.06656 & 0.78552 & +0.71896 \\
 & \texttt{8J48} (\texttt{A}--\texttt{B}) & Inter-species heteromeric & 12,417 & 158 & 0.44776 & 0.94852 & +0.50076 \\
 & \texttt{7YUJ} (\texttt{A}--\texttt{B}) & Homomer-like & 1,517 & 327 & 0.09496 & 0.43584 & +0.34088 \\
 & \texttt{8PR6} (\texttt{A}--\texttt{B}) & Homomer-like & 1,493 & 43 & 0.05188 & 0.34216 & +0.29028 \\
 & \texttt{8HHV} (\texttt{A}--\texttt{D}) & Homomer-like & 4,316 & 294 & 0.56732 & 0.83340 & +0.26608 \\
 & \texttt{8Q00} (\texttt{A}--\texttt{B}) & Inter-species heteromeric & 7,971 & 86 & 0.67020 & 0.85044 & +0.18024 \\
 & \texttt{8J49} (\texttt{A}--\texttt{B}) & Inter-species heteromeric & 4,220 & 505 & 0.13092 & 0.28416 & +0.15324 \\
 & \texttt{8ABT} (\texttt{A}--\texttt{A-2}) & Homomer-like & 3,938 & 151 & 0.81004 & 0.94712 & +0.13708 \\
\midrule
\multirow{4}{*}{FBv2} & \texttt{7FFN} (\texttt{C}--\texttt{D}) & Inter-species heteromeric & 15,566 & 212 & 0.12208 & 0.63104 & +0.50896 \\
 & \texttt{9HIS} (\texttt{B}--\texttt{D}) & Intra-species heteromeric & 15,114 & 196 & 0.63196 & 0.81644 & +0.18448 \\
 & \texttt{9O5W} (\texttt{A}--\texttt{B}) & Homomer-like & 2,135 & 56 & 0.74464 & 0.90940 & +0.16476 \\
 & \texttt{8CID} (\texttt{A}--\texttt{B}) & Homomer-like & 1,365 & 116 & 0.65676 & 0.79528 & +0.13852 \\
\midrule
\multirow{12}{*}{HD439} & \texttt{7ZHR\_1} (\texttt{A}--\texttt{B}) & Intra-species heteromeric & 16,384 & 330 & 0.10476 & 0.86108 & +0.75632 \\
 & \texttt{8BYP\_1} (\texttt{A}--\texttt{B}) & Intra-species heteromeric & 8,192 & 126 & 0.00768 & 0.56012 & +0.55244 \\
 & \texttt{8FZZ\_2} (\texttt{A}--\texttt{B}) & Intra-species heteromeric & 1,609 & 68 & 0.30708 & 0.80464 & +0.49756 \\
 & \texttt{8GXE\_1} (\texttt{A}--\texttt{B}) & Intra-species heteromeric & 14,818 & 218 & 0.31168 & 0.72912 & +0.41744 \\
 & \texttt{7ACX\_1} (\texttt{A}--\texttt{B}) & Intra-species heteromeric & 6,835 & 117 & 0.25400 & 0.60648 & +0.35248 \\
 & \texttt{7WDK\_1} (\texttt{A}--\texttt{B}) & Inter-species heteromeric & 3,170 & 407 & 0.29568 & 0.60808 & +0.31240 \\
 & \texttt{7Y8V\_1} (\texttt{A}--\texttt{B}) & Intra-species heteromeric & 8,192 & 131 & 0.20072 & 0.44360 & +0.24288 \\
 & \texttt{7P99\_1} (\texttt{A}--\texttt{B}) & Inter-species heteromeric & 8,752 & 127 & 0.60416 & 0.80168 & +0.19752 \\
 & \texttt{7NA9\_1} (\texttt{A}--\texttt{B}) & Inter-species heteromeric & 10,074 & 170 & 0.01984 & 0.18612 & +0.16628 \\
 & \texttt{7TCX\_1} (\texttt{A}--\texttt{B}) & Intra-species heteromeric & 2,666 & 483 & 0.69208 & 0.85764 & +0.16556 \\
 & \texttt{7MNX\_1} (\texttt{A}--\texttt{B}) & Intra-species heteromeric & 13,121 & 349 & 0.62940 & 0.78488 & +0.15548 \\
 & \texttt{8AJY\_1} (\texttt{A}--\texttt{B}) & Intra-species heteromeric & 1,454 & 114 & 0.73808 & 0.86092 & +0.12284 \\
\bottomrule
\end{tabular}%
}
\end{table*}


\begin{table*}[t]
\centering
\small
\caption{Illustrative exploration and final-selection outcomes across three benchmarks. Default denotes AF3-Default; Submission oracle is the retrospectively best tested submission (including Default); Selection is the original final Agent choice. Each score is the mean DockQ over 25 structures for the indicated interface; parentheses give the submission index. Each target shown has one evaluated interface. `No improvement' means that none of the tested alternatives exceeded Default; it does not refer to untested programs. FoldBench chain IDs refer to assembly 1; HD439 labels follow the released benchmark structures. These selected examples do not estimate outcome frequencies.}
\label{tab:exploration-selection-cases-all}
\resizebox{\textwidth}{!}{%
\begin{tabular}{lllrrr}
\toprule
Dataset & Interface & Type & Default & Submission oracle & Selection \\
\midrule
\multicolumn{6}{l}{\textit{a) Improvement found and selected}} \\
FBv1 & 9C5S (A--B) & Homomer-like & 0.07672 (1) & 0.62048 (2) & 0.62048 (2) \\
FBv1 & 8J48 (A--B) & Inter-species heteromeric & 0.44776 (1) & 0.94852 (3) & 0.94852 (3) \\
FBv2 & 9OC6 (A--C) & Inter-species heteromeric & 0.18380 (1) & 0.90320 (7) & 0.90320 (7) \\
FBv2 & 8WVY (A--B) & Intra-species heteromeric & 0.03564 (1) & 0.75060 (6) & 0.75060 (6) \\
HD439 & 7PPZ\_1 (A--B) & Inter-species heteromeric & 0.04400 (1) & 0.85116 (5) & 0.85116 (5) \\
HD439 & 7E5B\_1 (A--B) & Inter-species heteromeric & 0.17972 (1) & 0.92192 (5) & 0.92192 (5) \\
\midrule
\multicolumn{6}{l}{\textit{b) Improvement found but missed}} \\
FBv1 & 8BZ2 (A--B) & Homomer-like & 0.03264 (1) & 0.64628 (2) & 0.03264 (1) \\
FBv2 & 7Z73 (A--F) & Inter-species heteromeric & 0.30144 (1) & 0.58060 (4) & 0.05812 (7) \\
FBv2 & 9KL2 (A--B) & Homomer-like & 0.38404 (1) & 0.79248 (5) & 0.38404 (1) \\
FBv2 & 7XOZ (A--D) & Homomer-like & 0.24872 (1) & 0.41188 (3) & 0.04768 (8) \\
HD439 & 8JC5\_1 (A--B) & Intra-species heteromeric & 0.00800 (1) & 0.58084 (2) & 0.00800 (5) \\
HD439 & 7V8E\_1 (A--B) & Inter-species heteromeric & 0.09420 (1) & 0.66156 (2) & 0.09420 (1) \\
\midrule
\multicolumn{6}{l}{\textit{c) No improvement; Default retained}} \\
FBv1 & 8AND (A--B) & Homomer-like & 0.52344 (1) & 0.52344 (1) & 0.52344 (1) \\
FBv1 & 8UPB (A--A-2) & Homomer-like & 0.92020 (1) & 0.92020 (1) & 0.92020 (1) \\
FBv2 & 9GWJ (C--D) & Intra-species heteromeric & 0.74288 (1) & 0.74288 (1) & 0.74288 (1) \\
FBv2 & 9E55 (B--C) & Inter-species heteromeric & 0.56548 (1) & 0.56548 (1) & 0.56548 (1) \\
HD439 & 8H4X\_1 (A--B) & Intra-species heteromeric & 0.91204 (1) & 0.91204 (1) & 0.91204 (1) \\
HD439 & 8GU7\_2 (A--B) & Inter-species heteromeric & 0.89944 (1) & 0.89944 (1) & 0.89944 (1) \\
\midrule
\multicolumn{6}{l}{\textit{d) No improvement; selection regressed}} \\
FBv1 & 8GQP (A--B) & Intra-species heteromeric & 0.51828 (1) & 0.51828 (1) & 0.23116 (3) \\
FBv2 & 9KO4 (A--B) & Intra-species heteromeric & 0.67592 (1) & 0.67592 (1) & 0.56896 (2) \\
FBv2 & 9LIL (A--B) & Homomer-like & 0.75012 (1) & 0.75012 (1) & 0.64836 (5) \\
HD439 & 9BQ2\_2 (A--B) & Intra-species heteromeric & 0.16796 (1) & 0.16796 (1) & 0.02080 (7) \\
HD439 & 8GBQ\_1 (A--B) & Intra-species heteromeric & 0.89784 (1) & 0.89784 (1) & 0.76972 (7) \\
HD439 & 8Q0N\_1 (A--B) & Intra-species heteromeric & 0.40504 (1) & 0.40504 (1) & 0.29432 (4) \\
\bottomrule
\end{tabular}}
\end{table*}

  \begin{figure}[!htb]
      \centering
    \includegraphics[width=\linewidth]{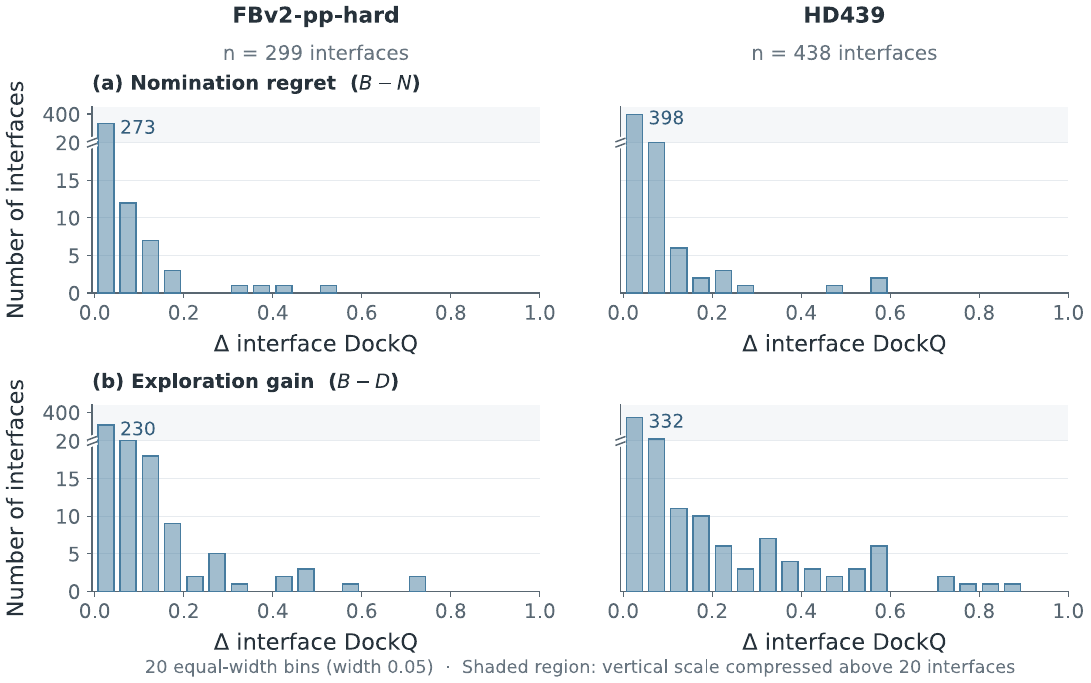}
      \caption{
\textbf{Exploration and nomination under predictor-derived and ground-truth feedback.} Columns correspond to the three benchmarks, and semi-transparent histograms overlay the standard predictor-feedback setting and the ground-truth-guided setting.
}
      \label{fig:exploration-selection_2hd}
  \end{figure}
\section{Seeds Ablation}
\label{seeds}
We assess sensitivity to the AF3 seeds by evaluating the fixed Agent
nominations and Default on FoldBench-v1 (238 targets and 278 interfaces).
The original seeds are \texttt{42, 66, 101, 2024, 8888}; the fresh, disjoint
seed set is \texttt{137, 271, 503, 1009, 2027}. Each setting uses five
diffusion samples per seed and ten recycles, yielding 25 structures per target.
For each configuration, only the model seeds change; its MSA content and
row order, templates, other input fields, and MSA overlap setting are preserved.
The agent does not repeat exploration or nomination. For the 40 targets whose
nomination is exactly Default, the same completed fresh-seed prediction and
evaluation supplies both scores; the remaining 198 Default configurations
are separately re-predicted and evaluated.
\begin{table*}[t]
\centering
\caption{
Seed robustness of fixed Agent nominations on FoldBench-v1.
Results report mean DockQ over interfaces; $\Delta$ denotes Agent minus Default.
}
\label{tab:fb1_seed_ablation}

\small
\setlength{\tabcolsep}{4.2pt}
\renewcommand{\arraystretch}{1.12}

\begin{tabular}{@{}l c rrrr rrrr@{}}
\toprule
& 
& \multicolumn{4}{c}{Original seeds}
& \multicolumn{4}{c}{Fresh seeds} \\
\cmidrule(lr){3-6}
\cmidrule(lr){7-10}

Interface group
& $N$
& Default
& Agent
& $\Delta$
& $p$
& Default
& Agent
& $\Delta$
& $p$ \\
\midrule

Homomeric
& 194
& 0.539
& 0.554
& +0.014
& 0.003
& 0.538
& 0.553
& +0.016
& 0.010 \\

Intra-species heteromeric
& 65
& 0.751
& 0.752
& +0.001
& 0.873
& 0.722
& 0.731
& +0.008
& 0.484 \\

Inter-species heteromeric
& 19
& 0.465
& 0.612
& +0.147
& 0.030
& 0.449
& 0.606
& +0.157
& 0.036 \\

\midrule
\textbf{All}
& \textbf{278}
& \textbf{0.584}
& \textbf{0.604}
& \textbf{+0.020}
& \textbf{$<0.001$}
& \textbf{0.575}
& \textbf{0.598}
& \textbf{+0.024}
& \textbf{0.001} \\
\bottomrule
\end{tabular}

\vspace{2pt}
\begin{minipage}{0.96\textwidth}
\footnotesize
DockQ is first averaged over 25 structures per interface and then
equally over interfaces within each group.
$p$-values are unadjusted two-sided paired $t$-tests on interface-level
Agent--Default differences. The overall row contains all 278 interfaces.
\end{minipage}

\end{table*}

As shown in Table~\ref{tab:fb1_seed_ablation},
the overall mean Agent--Default difference is +0.020
under the original seeds ($p<0.001$) and
+0.024 under the fresh seeds ($p=0.001$).
The tests use interface means as paired observations, treating interfaces as
independent; the 25 structures are not counted as independent replicates.
This experiment measures fixed nominations under one additional seed set.

\FloatBarrier
\section{Snapshot of workspace folder}
\label{workspace}
Each episode runs in a private sandbox containing its task files and a
persistent agent-writable working directory, \texttt{/workspace/}. We refer
to the complete sandbox file environment as the episode workspace. The
runtime supplies the initial target input, submission records, and prediction
outputs, while the agent uses \texttt{analysis} to inspect these files and
save its own programs, derived statistics, notes, and candidate MSA
organizations.

The directory layout for AF3-backed episodes is shown below. RO and RW denote
read-only and read--write access from within the sandbox, respectively; the
host runtime can still update the backing files. Iteration directories are
numbered \texttt{001}, \texttt{002}, and so on. Ground-truth-dependent
directories, marked GT-guided below, are absent from the Blind sandbox.

\begingroup
\lstdefinestyle{msaWorkspaceLayout}{
  language={},
  basicstyle=\footnotesize\ttfamily\color{black},
  backgroundcolor=\color{black!1},
  rulecolor=\color{black!22},
  columns=fullflexible,
  keepspaces=true,
  showstringspaces=false,
  breaklines=true,
  breakatwhitespace=true,
  numbers=none,
  frame=single,
  framerule=0.3pt,
  framesep=6pt,
  xleftmargin=6pt,
  xrightmargin=6pt,
  aboveskip=0.7\baselineskip,
  belowskip=0.7\baselineskip,
  lineskip=0.5pt
}

\begin{lstlisting}[style=msaWorkspaceLayout]
/
|-- input/                                      [RO]
|   `-- source.json
|-- observation/                                [RO]
|   `-- current.json
|-- artifacts/                                  [RO]
|   `-- <iteration>/
|       |-- builder.py
|       |-- candidate.json
|       |-- builder_stdout.txt
|       |-- builder_stderr.txt
|       |-- input.json
|       |-- run.log
|       `-- predictions/target/
|           |-- seed-<seed>_sample-<sample>/  (all seeds/samples)
|           |   |-- *_model.cif
|           |   |-- *_summary_confidences.json
|           |   `-- *_confidences.json
|           |-- target_ranking_scores.csv
|           |-- target_model.cif
|           |-- target_summary_confidences.json
|           |-- target_confidences.json
|           |-- target_data.json
|           `-- TERMS_OF_USE.md
|-- workspace/                                  [RW]
|   |-- analysis/<analysis_id>.py
|   |-- submissions/<iteration>/candidate.json
|   `-- ...  (agent-created files and directories)
|-- ground_truth/                               [RO, GT-guided]
|   |-- evaluation.json
|   `-- raw/
|-- evaluations/                                [RO, GT-guided]
|   `-- <iteration>/
|       `-- ...  (per-sample scores, reports, logs)
`-- tmp/                                        [RW, per-call]
\end{lstlisting}
\endgroup

\paragraph{Inputs and submission history.}
\texttt{/input/source.json} contains the complete precomputed AF3 input,
including query sequences, the original \texttt{pairedMsa} and
\texttt{unpairedMsa} strings, supplied templates, and other original input
fields. Only its top-level \texttt{name} is replaced by \texttt{target}.
It is available before the first submission and remains unchanged throughout
the episode.

\texttt{/observation/current.json} records the remaining evaluation budget
and the history of successful submissions. Each entry contains the candidate
identifier, strategy, experimental summary, builder and artifact paths, and
profile-visible feedback. The history is initially empty; the runtime updates
this file after each successful submission.

\paragraph{Submission artifacts.}
Each \texttt{/artifacts/<iteration>/} directory stores the executed builder,
its generated candidate and logs, the actual AF3 input for that submission,
and the prediction outputs. The latter include structures, confidence
summaries, and full confidence arrays such as PAE, pLDDT, and contact
probabilities for each seed and sample.

Under the five-seed, five-sample setting, the seed/sample pattern expands to
25 directories; all 25 structures and their confidence files are retained.
At the \texttt{predictions/target/} root,
\texttt{target\_ranking\_scores.csv} records the samples' ranking scores.
The root-level \texttt{target\_model.cif},
\texttt{target\_summary\_confidences.json}, and
\texttt{target\_confidences.json} additionally contain the structure and
confidence outputs of the sample with the highest AF3 ranking score. The
agent can inspect every sample in its corresponding subdirectory, and
artifacts from earlier iterations remain available for comparison.

\paragraph{Agent-created files.}
Under \texttt{/workspace/}, the runtime automatically saves each
\texttt{analysis} program in \texttt{analysis/} and allocates
\texttt{submissions/<iteration>/candidate.json} as the builder's output path,
passed through \texttt{--output}. The builder writes this file; the runtime
validates it and copies it into the corresponding iteration artifact
directory.

The agent may choose any builder source path within \texttt{/workspace/} and
organize its remaining files freely, for example as \texttt{builders/},
\texttt{notes.md}, helper modules, or statistical tables. These example names
are not required by the runtime. During submission, a read-only copy of the
chosen script is temporarily mounted as \texttt{/builder/builder.py} for
execution.

\paragraph{Ground-truth visibility.}
In GT-guided episodes, \texttt{/ground\_truth/evaluation.json} describes the
target's evaluation interfaces and the location of the reference files
mounted under \texttt{/ground\_truth/raw/}.
\texttt{/evaluations/<iteration>/} exposes the corresponding detailed
evaluation outputs, including per-sample metric tables, raw evaluator reports,
and logs.

Blind episodes retain evaluation results on the host but expose neither
directory; their submission feedback contains only model-derived signals and
artifact paths. Evaluator outputs are stored separately from
\texttt{/artifacts/}.

\paragraph{Persistence and context.}
The episode files remain available across tool calls and round-context resets.
Each \texttt{analysis} call and builder execution starts a fresh Python
process with \texttt{/workspace/} as its working directory;
\texttt{/tmp/} is recreated for each execution and discarded afterward.

File contents are not automatically inserted into the model context: the agent
inspects them through \texttt{analysis} and receives the program's standard
output and error as tool results. Compact submission feedback is also returned
directly by \texttt{submit}.

\section{GT-Guide Results}
\label{gtgr}
\paragraph{Setting and evaluation.}
We compare the standard predictor-feedback setting (Blind) with a
ground-truth-guided setting on FBv1-pp. The GT-guided agent can inspect
reference structures, DockQ scores, and interface-level evaluation
reports during exploration; these resources remain hidden from Blind.
The agent decides when to terminate, while the GT-guided
\texttt{finish} tool automatically selects the successfully tested
submission with the highest target-level mean DockQ, including Default.
This guarantees zero target-level nomination regret, but
interface-level nomination regret may remain nonzero.
Both settings use the same AF3 inference configuration: five seeds,
five diffusion samples per seed, and ten recycles. The episode limits
are configured as 60 successful submissions and 600 LLM calls, with early termination
permitted. Submission counts include the initial Default evaluation.

All 239 GT-guided episodes completed. We report comparisons on the
238 targets and 278 evaluated interfaces shared with Blind; 8BVK has
no completed Blind result and is excluded. Each interface score is
the mean DockQ over 25 AF3 structures. A proposal's target-level score
averages these interface means equally within the target. For each
interface, the best-tested score is the highest mean DockQ across all
successfully tested submissions, including Default. This retrospective
interface-wise oracle may select different submissions for different
interfaces of the same target and does not replace either agent's
actual final nomination. Interface-level exploration gain is the
best-tested score minus the Default score for that interface;
nomination regret is the best-tested score minus the score of the
actual nominated submission on the same interface.

\paragraph{Overall performance.}
Table~\ref{tab:gt-guide-overall} reports results using equal weighting
over evaluated interfaces, consistent with the main benchmark tables.
GT-guided achieves mean DockQ 0.6425, compared with 0.6041 for Blind's
final nominations and 0.6183 for its interface-wise best-tested
scores. The improvement over Blind's interface-wise oracle is
0.0242 DockQ overall, with positive mean differences in all three
interface groups. GT-guided final submissions therefore outperform
Blind's retrospective interface-wise oracle on average, indicating
improvements beyond those discovered during Blind exploration.

\begin{table}[t]
\centering
\caption{Completed GT-guided and Blind results on FBv1-pp.
DockQ is averaged over 25 structures per interface and then equally
over interfaces within each group. Blind nomination is the original
agent choice; Blind best tested is the retrospective interface-wise
submission oracle. GT-guided reports its actual final submission,
automatically selected by target-level mean DockQ.
$\Delta_{\mathrm{best}}$ is GT-guided minus Blind best
tested. All methods use the same 278 interfaces from 238 targets.}
\label{tab:gt-guide-overall}
\small
\setlength{\tabcolsep}{4pt}
\renewcommand{\arraystretch}{1.12}
\begin{tabular}{@{}lrrrrrr@{}}
\toprule
Interface group & $n_{\mathrm{int}}$ & Default
& \shortstack{Blind\\nomination} & \shortstack{Blind\\best tested}
& GT-guided & $\Delta_{\mathrm{best}}$ \\
\midrule
Homomeric & 194 & 0.5393 & 0.5537 & 0.5661 & 0.5900 & +0.0239 \\
Same-species heteromeric & 65 & 0.7509 & 0.7521 & 0.7730 & 0.7937 & +0.0206 \\
Cross-species heteromeric & 19 & 0.4649 & 0.6123 & 0.6215 & 0.6606 & +0.0391 \\
\midrule
\textbf{All} & \textbf{278} & \textbf{0.5837} & \textbf{0.6041}
& \textbf{0.6183} & \textbf{0.6425} & \textbf{+0.0242} \\
\bottomrule
\end{tabular}
\end{table}

\paragraph{Longer exploration trajectories.}
Across the 238 matched targets, GT-guided makes 2,613 successful
submissions, compared with 1,558 for Blind. The mean episode length
increases from 6.55 to 10.98 submissions, a 67.7\% increase, and the
median increases from 6 to 8. GT-guided makes more submissions on
152 of 238 targets (63.9\%). With interfaces weighted equally, mean
exploration gain increases from 0.0346 to 0.0594 DockQ (71.7\%).
The interface-wise best-tested score under GT guidance exceeds its
Blind counterpart on 161 of 278 interfaces (57.9\%). These gains
use the retrospective interface-wise best-tested scores in both
settings; the GT-guided oracle averages 0.6431 DockQ, compared with
0.6425 for its actual final nominations. GT-guided has zero
interface-level nomination regret on 251 of 278 interfaces (90.3\%),
with a mean regret of 0.000593 DockQ across all interfaces.
Thus, longer exploration under ground-truth guidance is accompanied
by larger discovered improvements in the completed trajectories.

\begin{figure}[t]
\centering
 \includegraphics[width=\linewidth]{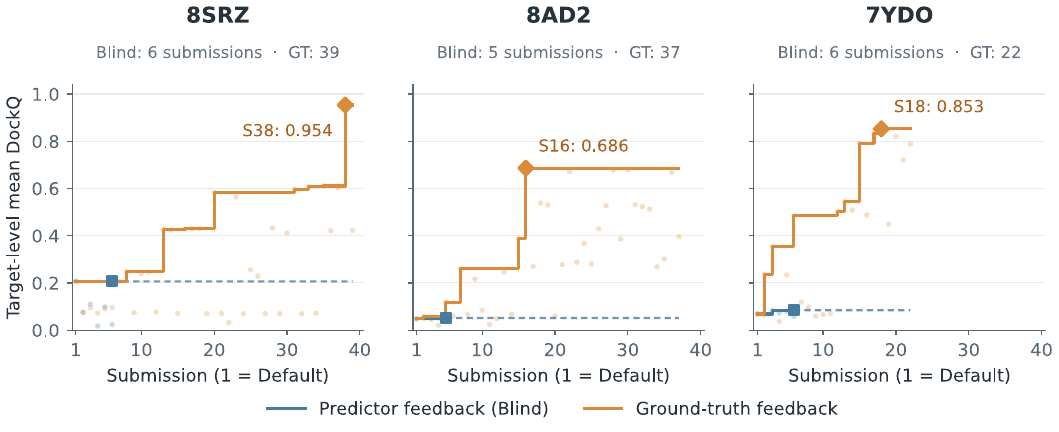}
\caption{Exploration trajectories for three FBv1-pp targets.
Faint dots show individual submission scores; solid step curves show
the best mean DockQ found so far. Each target has one evaluated
interface, so interface-level and target-level scores coincide.
Blue squares mark
the end of Blind exploration, and dashed blue extensions retain
Blind's best score as a reference after termination. Orange diamonds
mark the first occurrence of the best GT-guided submission.
Blind DockQ is evaluated retrospectively and was unavailable to the
agent during exploration and nomination.}
\label{fig:gt-guide-trajectories}
\end{figure}

\paragraph{Case comparisons.}
We examine three illustrative targets with large improvements over
Blind's best tested submission (Table~\ref{tab:gt-guide-cases} and
Figure~\ref{fig:gt-guide-trajectories}). Let $T_B$ denote the number
of submissions in the Blind episode. For each case, we also report
the best score within the first $T_B$ submissions of the GT-guided
episode itself, showing how much additional gain appears later in
that trajectory. All three cases have one evaluated benchmark
interface, and all scores average 25 AF3 structures per submission
generated with the same seed set.

\begin{table}[t]
\centering
\caption{Selected cases of extended GT-guided exploration.
Submission counts include Default as S1. The GT prefix column is the
best score among the first $T_B$ GT-guided submissions, where $T_B$
is the full Blind episode length. The last column gives the first
submission attaining the best score in the complete GT-guided episode.
These cases illustrate late gains and do not estimate their frequency
across the benchmark.}
\label{tab:gt-guide-cases}
\small
\setlength{\tabcolsep}{4pt}
\renewcommand{\arraystretch}{1.12}
\begin{tabular}{@{}lrrrrrrr@{}}
\toprule
Target & \shortstack{Submissions\\Blind / GT} & Default
& \shortstack{Blind\\nomination} & \shortstack{Blind\\best tested}
& \shortstack{GT best\\first $T_B$} & \shortstack{GT best\\full episode}
& \shortstack{Best GT\\submission} \\
\midrule
8SRZ & 6 / 39 & 0.2068 & 0.2068 & 0.2068 & 0.2068 & 0.9535 & S38 \\
8AD2 & 5 / 37 & 0.0477 & 0.0514 & 0.0514 & 0.1180 & 0.6858 & S16 \\
7YDO & 6 / 22 & 0.0687 & 0.0666 & 0.0844 & 0.4838 & 0.8525 & S18 \\
\bottomrule
\end{tabular}
\end{table}

\paragraph{8SRZ: improvement after an unproductive initial search.}
Blind ends after six submissions and retains Default, with mean DockQ
0.2068 on the evaluated A--B interface. The first six GT-guided
submissions likewise fail to improve on Default. Continued GT-guided
exploration removes three fragmentary source MSA rows at S8, reaching
0.2467, and then explores the composition and multiplicities of the
remaining homolog rows. Mean DockQ reaches 0.4276 at S13 and 0.5839
at S20. S38, a nine-record alignment per chain with revised homolog
weighting, achieves 0.9535 and is retained after one further trial.
The full GT-guided episode contains 39 submissions. Its entire
0.7468 advantage over Blind's best tested score is discovered after
the first six GT-guided submissions.

\paragraph{8AD2: continued refinement of interface-informed MSA weighting.}
Blind tests five submissions, attaining its best and nominated score
of 0.0514 at S3 on the evaluated A--B interface. GT-guided reaches
0.1180 within its first five submissions and continues exploring.
At S7, it filters source rows for coverage of the reference interface
residues, reaching 0.2608 with 35 non-query rows. Later proposals
change the relative weight of rows that differ at those positions.
S16 supplies four copies of each of five divergent rows while keeping
the other 30 non-query rows once, producing 51 records per chain
including the query and achieving 0.6858. GT-guided completes 37
submissions and retains S16. Exploration beyond its first five
submissions contributes a further 0.5678 DockQ over its own prefix best.

\paragraph{7YDO: refining homolog choice and assignment across chains.}
Blind makes six submissions on this eight-chain target, with a best
tested score of 0.0844 and a nominated score of 0.0666 on the evaluated
A--A-2 interface. GT-guided explores shallow alignments containing
the query plus one homolog per chain, with chain-specific homolog
assignments, reaching 0.4838 at S6. Further exploration tests new
homolog sets and their assignments across the eight chains. S15
reaches 0.7911; reassigning this homolog set yields 0.8335 at S17 and
0.8525 at S18. The episode ends after 22 submissions and retains S18.
The later refinement adds 0.3687 DockQ beyond the best score in the
first six GT-guided submissions, yielding a final advantage of 0.7681
over Blind's best tested proposal.

\section{Case Study Information}
\label{cs}

\subsection{FBv1: 8P9C}
\label{app:case_8p9c}
\paragraph{Target and original input.}
8P9C comprises the tetramerization domains of p63 and p73 in complex
with the designed ankyrin repeat protein DARPin 1810
F11~\cite{Strubel2023DARPins}.
The target contains two copies of each protein sequence: p63 in
chains A/K, p73 in chains B/L, and DARPin in chains C/M.
We evaluate the A--C and B--C interfaces.
Table~\ref{tab:8p9c_original_input} lists the protein lengths and MSA
depths in the original input JSON.
The two copies of each protein share the same MSA inputs.
The \texttt{pairedMsa} field supplies candidates for subsequent
pairing; its depth is not the number of successfully paired rows.

\begin{table}[t]
  \caption{Original input statistics for 8P9C. Length is measured
  in amino acid residues per chain. Paired and Unpaired denote
  the depths of the \texttt{pairedMsa} and \texttt{unpairedMsa}
  fields, respectively. Each depth includes one query row and is
  shared by the two identical copies, rather than summed over them.}
  \label{tab:8p9c_original_input}
  \centering
  \small
  \setlength{\tabcolsep}{4pt}
  \renewcommand{\arraystretch}{1.1}
  \begin{tabular}{@{}lcrrr@{}}
    \toprule
    \textbf{Protein} & \textbf{Chains} & \textbf{Length}
    & \textbf{Paired} & \textbf{Unpaired} \\
    \midrule
    p63    & A/K & 61  & 2,378  & 428    \\
    p73    & B/L & 50  & 2,473  & 303    \\
    DARPin & C/M & 159 & 50,000 & 16,639 \\
    \bottomrule
  \end{tabular}
\end{table}

\paragraph{Default MSA construction.}
The default AF3 pipeline matches sequences using species identifiers
and removes unpaired sequences already represented in the retained
paired block.
This overlap removal reduces the unpaired depths of p63, p73, and
DARPin to 174, 113, and 13,872, respectively.
The DARPin unpaired alignment is subsequently cropped to 8,192 rows;
the p63 and p73 unpaired alignments require no further cropping.
The shorter unpaired alignments are padded with gap rows before
concatenation across chains.
The paired and unpaired blocks are then stacked, producing a
featurized MSA of 16,384 rows: 8,192 in each block, with one query
row in the paired block.
Throughout this case study, featurized depth refers to the merged
MSA produced by preprocessing, before any model-internal MSA sampling.

Table~\ref{tab:8p9c_default_layout} reports the resulting row occupancy.
Excluding the query, the paired block contains 2,297 rows spanning
all three protein types, 38 p63--p73 rows, 7 p73--DARPin rows,
20 p73-only rows, and 5,829 DARPin-only rows.
A row covering only one protein type can still cover both copies
of that protein, such as C and M.
Across the two blocks, 13,847 rows (84.5\%) contain sequence
information only for the DARPin chains.
The total non-gap depths for p63, p73, and DARPin are 2,510, 2,476,
and 16,326, respectively.

\begin{table}[t]
  \caption{Row occupancy after default AF3 featurization.
  Each protein type represents both identical copies.
  A plus sign denotes a non-gap sequence and a minus sign denotes
  an all-gap sequence. The 2,298 paired rows covering all three
  types include one query row. Co-occupancy in the unpaired block
  results from concatenating independent alignments and does not
  indicate species matching.}
  \label{tab:8p9c_default_layout}
  \centering
  \small
  \setlength{\tabcolsep}{4pt}
  \renewcommand{\arraystretch}{1.1}
  \begin{tabular}{@{}cccrr@{}}
    \toprule
    \textbf{p63} & \textbf{p73} & \textbf{DARPin}
    & \textbf{Paired} & \textbf{Unpaired} \\
    \midrule
    $+$ & $+$ & $+$ & 2,298 & 113   \\
    $+$ & $+$ & $-$ & 38    & 0     \\
    $+$ & $-$ & $+$ & 0     & 61    \\
    $-$ & $+$ & $+$ & 7     & 0     \\
    $-$ & $+$ & $-$ & 20    & 0     \\
    $-$ & $-$ & $+$ & 5,829 & 8,018 \\
    \midrule
    \multicolumn{3}{@{}l}{\textbf{Total}}
    & \textbf{8,192} & \textbf{8,192} \\
    \bottomrule
  \end{tabular}
\end{table}

\paragraph{Prediction and evaluation protocol.}
Each submission uses five prediction seeds (42, 66, 101, 2024, and
8888), with five diffusion samples per seed, yielding 25 predictions.
For each evaluated interface, we report DockQ averaged over these
25 predictions; the case-level mean is the arithmetic mean of the
A--C and B--C interface scores.
Mean ipTM is also averaged over the 25 predictions.
The agent operates in the blind setting, with access to prediction
confidence but not the DockQ evaluation scores.
The DockQ values reported here are obtained by subsequent evaluation.
For the default submission, the A--C and B--C mean DockQ scores are
0.0666 and 0.0294, respectively, giving a case-level mean of
0.0480; the mean ipTM is 0.310.

\paragraph{Agent construction and submission trajectory.}
The agent constructs MSAs with explicitly specified row
correspondence.
It uses relative sequence identity to the p63 and p73 queries to
reduce mixing between paralogous candidates, forms p63--p73 pairs
using species or taxonomy identifiers, and adds separate
supplemental blocks for each protein.
Columns for proteins not participating in a row are filled with
gaps, and the same rows are used for the two copies of each protein.
The constructed alignments are supplied through
\texttt{unpairedMsa}, with \texttt{pairedMsa} empty and overlap
resolution disabled, preserving the specified row correspondence.
As for the default input, depths below refer to the featurized MSA.

Table~\ref{tab:8p9c_submissions} summarizes all six submissions.
Submission 2 retains 180 p63--p73 pairs without additional
supplemental rows and achieves a mean DockQ of 0.0391.
Submission 3 expands the pair set to 188 rows and adds 80 p63,
104 p73, and 150 DARPin supplemental rows, increasing mean DockQ
to 0.869.
Because both the paired and supplemental blocks change between
these submissions, this comparison does not isolate the effect
of any single block.

The agent selects submission 4 using the available confidence
signals.
Its construction contains 202 p63--p73 paired rows, consisting of
188 pairs from the UniProt candidate pools and 14 additional pairs
from UniRef90 candidates, together with 26 p63, 45 p73, and
150 DARPin supplemental rows.
The DARPin supplement is selected directly from its original
\texttt{unpairedMsa}: candidates must cover at least 70\% of the
query and have at least 25\% sequence identity over aligned
non-gap positions.
Candidates are considered in descending order of query identity;
a candidate is skipped if its identity to any already selected
sequence is at least 85\% over their overlapping non-gap positions.
Selection stops at the prescribed supplement size.
The selected construction yields a featurized MSA with
$202 + 26 + 45 + 150 + 2 = 425$ rows, including two query rows.
Table~\ref{tab:8p9c_agent_layout} shows the resulting row occupancy.
The corresponding non-gap depths are 230, 249, and 152 for
p63, p73, and DARPin, respectively, including the query rows.
Its mean ipTM is 0.839, the highest among the six submissions.
Subsequent evaluation yields mean DockQ scores of 0.786 for A--C
and 0.946 for B--C, giving a case-level mean of 0.866.
Submission 3 has a slightly higher subsequent mean DockQ
(0.869), so submission 4 is the agent-selected construction,
rather than the construction with the highest evaluation score.

\begin{table}[t]
  \caption{Row occupancy after AF3 featurization of the
  agent-selected construction (submission 4).
  As in Table~\ref{tab:8p9c_default_layout}, a plus sign denotes
  a non-gap sequence and a minus sign denotes an all-gap sequence.
  The p63, p73, and DARPin columns represent chains A/K, B/L, and
  C/M, respectively. The two query rows comprise the query in
  the submitted alignment and the query that AF3 uses to
  initialize the empty paired block.}
  \label{tab:8p9c_agent_layout}
  \centering
  \small
  \setlength{\tabcolsep}{4pt}
  \renewcommand{\arraystretch}{1.1}
  \begin{tabular}{@{}lcccr@{}}
    \toprule
    \textbf{Block} & \textbf{p63} & \textbf{p73}
    & \textbf{DARPin} & \textbf{Rows} \\
    \midrule
    Query rows        & $+$ & $+$ & $+$ & 2   \\
    p63--p73 pairs    & $+$ & $+$ & $-$ & 202 \\
    p63 supplement    & $+$ & $-$ & $-$ & 26  \\
    p73 supplement    & $-$ & $+$ & $-$ & 45  \\
    DARPin supplement & $-$ & $-$ & $+$ & 150 \\
    \midrule
    \multicolumn{4}{@{}l}{\textbf{Total}}
    & \textbf{425} \\
    \bottomrule
  \end{tabular}
\end{table}

\begin{table*}[t]
  \caption{MSA constructions and results for all 8P9C submissions.
  A, B, and C denote the p63, p73, and DARPin protein groups,
  respectively, including their identical copies.
  A--B counts explicitly paired rows; A-only, B-only, and C-only
  count the separate supplemental rows. These four counts exclude
  query rows. Depth is the total featurized MSA depth, including
  one query row for Default and two for each manual construction.
  Native row occupancy and the selected agent construction are
  given in Tables~\ref{tab:8p9c_default_layout}
  and~\ref{tab:8p9c_agent_layout}, respectively.
  Scores are rounded to three significant figures.
  Bold scores indicate the highest value in each metric column,
  determined before rounding; the selected submission label is
  also bold. Arrows indicate that higher scores are better.
  DockQ was unavailable to the agent during its search.}
  \label{tab:8p9c_submissions}
  \centering
  \small
  \setlength{\tabcolsep}{3pt}
  \renewcommand{\arraystretch}{1.1}
  \begin{tabular}{@{}lrrrrrrrrr@{}}
    \toprule
    & \multicolumn{5}{c}{\textbf{MSA construction}}
    & \multicolumn{3}{c}{\textbf{DockQ} $\uparrow$}
    & \textbf{ipTM} $\uparrow$ \\
    \cmidrule(lr){2-6}\cmidrule(lr){7-9}\cmidrule(l){10-10}
    \textbf{Submission} & \textbf{A--B} & \textbf{A-only}
    & \textbf{B-only} & \textbf{C-only} & \textbf{Depth}
    & \textbf{A--C} & \textbf{B--C} & \textbf{Mean} & \textbf{Mean} \\
    \midrule
    1 (Default) & \multicolumn{4}{c}{Native construction} & 16,384
    & 0.0666 & 0.0294 & 0.0480 & 0.310 \\
    \midrule
    2 & 180 & 0  & 0   & 0   & 182
    & 0.0488 & 0.0294 & 0.0391 & 0.323 \\
    3 & 188 & 80 & 104 & 150 & 524
    & \textbf{0.786} & \textbf{0.951} & \textbf{0.869} & 0.822 \\
    \addlinespace
    \textbf{4 (selected)} & 202 & 26 & 45 & 150 & 425
    & 0.786 & 0.946 & 0.866 & \textbf{0.839} \\
    5 & 202 & 26 & 45 & 300 & 575
    & 0.202 & 0.231 & 0.216 & 0.563 \\
    6 & 202 & 26 & 45 & 100 & 375
    & 0.641 & 0.766 & 0.703 & 0.788 \\
    \bottomrule
  \end{tabular}
\end{table*}

\paragraph{Depth comparison within the agent search.}
Submissions 4--6 vary the DARPin supplement while keeping the
202 p63--p73 paired rows, the 26 p63 and 45 p73 supplemental rows,
and the other input settings, including templates and prediction
seeds, unchanged.
The 100-, 150-, and 300-row DARPin supplements are nested subsets
of the same ordered sequence list.
These budgets yield featurized depths of 375, 425, and 575 and
mean DockQ scores of 0.703, 0.866, and 0.216, respectively;
the corresponding mean ipTM values are 0.788, 0.839, and 0.563.
Thus, 150 supplemental rows perform best among the tested budgets,
and expanding the supplement to 300 rows substantially reduces
both prediction confidence and evaluated interface accuracy.
This comparison demonstrates sensitivity to expanding the selected
DARPin sequence set, which changes both its depth and composition.
It does not establish a universal optimum at 150 rows or separately
identify the effects of sequence count and sequence composition.
Submission 2 is not included as a zero-supplement condition in this
controlled comparison, because its p63--p73 pair set and its
p63/p73 supplements also differ.

\subsection{HD439: 7R86\_1}
\label{app:case_7r86_1}
\paragraph{Target and original input.}
7R86 captures an extracellular complex between the mouse Nogo
receptor RTN4R (NgR1) and the third thrombospondin type 1 repeat
(TSR3) of the adhesion G-protein-coupled receptor BAI1
(ADGRB1)~\cite{Wang2021RTN4}.
This interaction participates in a cell-surface recognition pathway
involved in neuronal development, with glycosylated TSR3 residues
contributing to the binding interface.
The benchmark target contains one copy of each protein fragment:
the RTN4R leucine-rich repeat domain in chain A (285 residues)
and BAI1 TSR3 in chain B (54 residues).
Chain labels refer to the benchmark input.
We evaluate the A--B interface.

Table~\ref{tab:7r86_1_original_input} reports the original input
statistics.
The \texttt{pairedMsa} fields provide candidate pools for subsequent
pairing; their depths are not counts of successfully paired rows.
Chain B has a query-only \texttt{unpairedMsa}, despite having
20,659 rows in its \texttt{pairedMsa} candidate pool.

\begin{table}[t]
  \caption{Original input statistics for 7R86\_1.
  Length is measured in amino acid residues.
  Paired and Unpaired denote the depths of the \texttt{pairedMsa}
  and \texttt{unpairedMsa} fields, respectively.
  Each depth includes one query row.}
  \label{tab:7r86_1_original_input}
  \centering
  \small
  \setlength{\tabcolsep}{4pt}
  \renewcommand{\arraystretch}{1.1}
  \begin{tabular}{@{}lcrrr@{}}
    \toprule
    \textbf{Protein} & \textbf{Chain} & \textbf{Length}
    & \textbf{Paired} & \textbf{Unpaired} \\
    \midrule
    RTN4R     & A & 285 & 17,486 & 4,456 \\
    BAI1 TSR3 & B & 54  & 20,659 & 1     \\
    \bottomrule
  \end{tabular}
\end{table}

\paragraph{Default MSA construction.}
The default AF3 pipeline pairs sequences using species identifiers.
The two candidate pools share 604 identifiers.
With at most 600 rows per identifier, matching produces 8,247
non-query paired rows before the global cap is applied.
Including the query and retaining the first 8,192 rows leaves
8,191 non-query A--B pairs and one query row.

Overlap resolution removes the query from each unpaired alignment,
reducing their depths from 4,456 to 4,455 for A and from 1 to 0
for B.
No further unpaired cropping is required.
The B columns are padded with gaps when the unpaired alignments
are concatenated across chains.
Stacking the paired and unpaired blocks yields
$8{,}192 + 4{,}455 = 12{,}647$ rows.
Table~\ref{tab:7r86_1_default_layout} shows the resulting occupancy.
The non-gap depths are 12,647 for A and 8,192 for B.
Throughout this case study, featurized depth denotes the merged
MSA row count before tensor padding and model-internal MSA sampling.

\begin{table}[t]
  \caption{Row occupancy after default AF3 featurization.
  A plus sign denotes a non-gap sequence and a minus sign denotes
  an all-gap sequence.
  The 8,192 paired rows include one query row.
  The unpaired block contains only A-chain sequence information.}
  \label{tab:7r86_1_default_layout}
  \centering
  \small
  \setlength{\tabcolsep}{5pt}
  \renewcommand{\arraystretch}{1.1}
  \begin{tabular}{@{}ccrr@{}}
    \toprule
    \textbf{A: RTN4R} & \textbf{B: BAI1}
    & \textbf{Paired} & \textbf{Unpaired} \\
    \midrule
    $+$ & $+$ & 8,192 & 0     \\
    $+$ & $-$ & 0     & 4,455 \\
    $-$ & $+$ & 0     & 0     \\
    \midrule
    \multicolumn{2}{@{}l}{\textbf{Total}}
    & \textbf{8,192} & \textbf{4,455} \\
    \bottomrule
  \end{tabular}
\end{table}

\paragraph{Prediction and evaluation protocol.}
Each submission uses five prediction seeds (42, 66, 101, 2024,
and 8888), with five diffusion samples per seed, yielding
25 predictions.
We report the A--B DockQ score averaged over these predictions.
Mean ipTM, pTM, and ranking score are also averaged over the
25 predictions.
The agent operates in the blind setting, with access to prediction
confidence but not DockQ.
All DockQ values below are obtained by subsequent evaluation.
Default yields a mean DockQ of 0.0365 and a mean ipTM of 0.194.

\paragraph{Agent construction.}
The selected construction combines filtered A--B pairs with separate
single-chain supplements.
All candidates are drawn from the original input MSAs.
After removing lowercase A3M insertion characters, the agent computes
query identity $I_q(s)$ as the number of matching non-gap residues
divided by the full query length, and coverage $C_q(s)$ as the
number of non-gap residues divided by the full query length.

For pairing, A candidates must satisfy $I_q(s)\geq 0.55$ and
$C_q(s)\geq 0.85$, while B candidates must satisfy
$I_q(s)\geq 0.60$ and $C_q(s)\geq 0.92$.
Both pools come from the original \texttt{pairedMsa} fields,
excluding the query records.
For each shared species identifier, the highest-identity candidate
is selected from each chain, with ties resolved by input order.
Identifiers are processed in sorted order, and a pair is skipped
if either aligned sequence has already been used in a retained pair.
The retained pairs are then ordered by decreasing summed query
identity across A and B.
This procedure produces 78 pairs.
The identity and coverage filters serve as heuristics for selecting
target-like candidates.

The agent adds 400 A-only rows from A's original
\texttt{unpairedMsa}, together with 222 higher-identity and
250 moderate-identity B-only rows from B's original
\texttt{pairedMsa}.
The latter pool provides the TSR-family supplement used by the agent.
Every supplemental row contains sequence information for only one
chain; the other chain is filled entirely with gaps.
The resulting alignments are supplied through \texttt{unpairedMsa},
with \texttt{pairedMsa} empty and overlap resolution disabled,
preserving the specified row correspondence.

The selected construction has a featurized depth of
$78 + 400 + 222 + 250 + 2 = 952$ rows.
Table~\ref{tab:7r86_1_agent_layout} shows its layout.
Including the query rows, the non-gap depths are 480 for A and
552 for B.
Thus, the intervention changes the composition and placement of
the available B sequences: it provides 472 explicit B-only
supplemental rows while reducing B's total non-gap depth relative
to Default.

\begin{table}[t]
  \caption{Row occupancy after AF3 featurization of the
  agent-selected construction (submission 3).
  The two query rows comprise the query in the submitted alignment
  and the query that AF3 uses to initialize the empty paired block.
  All supplemental blocks fill the other chain with gaps.
  Higher-identity and family supplements refer to the B candidate
  pools defined in the text.}
  \label{tab:7r86_1_agent_layout}
  \centering
  \small
  \setlength{\tabcolsep}{4pt}
  \renewcommand{\arraystretch}{1.1}
  \begin{tabular}{@{}lccr@{}}
    \toprule
    \textbf{Block} & \textbf{A} & \textbf{B} & \textbf{Rows} \\
    \midrule
    Query rows                    & $+$ & $+$ & 2   \\
    A--B pairs                    & $+$ & $+$ & 78  \\
    A supplement                  & $+$ & $-$ & 400 \\
    B higher-identity supplement  & $-$ & $+$ & 222 \\
    B family supplement           & $-$ & $+$ & 250 \\
    \midrule
    \multicolumn{3}{@{}l}{\textbf{Total}} & \textbf{952} \\
    \bottomrule
  \end{tabular}
\end{table}

\paragraph{Single-chain sequence selection.}
Algorithm~\ref{alg:7r86_1_supplement_selection} specifies the greedy
selection used for each supplemental pool.
The A-only pool requires $C_q(s)\geq 0.85$ and $I_q(s)\geq 0.25$;
it excludes aligned sequences already used in the A--B pairs and
uses a redundancy threshold of 0.97 with a budget of 400.
The higher-identity B pool requires $C_q(s)\geq 0.92$ and
$I_q(s)\geq 0.60$, excludes B sequences already used in pairs,
and uses a redundancy threshold of 0.98 with a budget of 400;
222 rows survive selection.
The B family pool requires $C_q(s)\geq 0.92$ and
$0.30\leq I_q(s)<0.60$, and uses a redundancy threshold of 0.90
with a budget of 250.
All pools exclude the initial query record.
For redundancy filtering, $J(s,t)$ is the fraction of identical
residues over positions where both sequences are non-gap, defined
as zero when there are no overlapping non-gap positions.
This pairwise identity uses a different denominator from $I_q(s)$.

\begin{algorithm}[t]
  \caption{Greedy selection of single-chain supplements}
  \label{alg:7r86_1_supplement_selection}
  \small
  \begin{algorithmic}[1]
    \Require Filtered candidate pool $R$, query $q$,
      redundancy threshold $\tau$, and row budget $K$
    \Ensure Ordered list $S$ of at most $K$ supplemental sequences
    \State Sort $R$ by decreasing $I_q(s)$,
      preserving input order for ties
    \State $S \gets [\,]$
    \ForAll{$s\in R$, in sorted order}
      \If{$|S|=K$}
        \State \textbf{break}
      \EndIf
      \If{$J(s,t)<\tau$ for every $t\in S$}
        \State Append $s$ to $S$
      \EndIf
    \EndFor
    \State \Return $S$
  \end{algorithmic}
\end{algorithm}

\paragraph{Submission trajectory.}
Table~\ref{tab:7r86_1_submissions} summarizes all six submissions.
Submission 2 uses only the 78 filtered pairs and achieves a mean
DockQ of 0.106.
Submission 3 retains exactly the same paired sequences and adds
the A-only and B-only supplements described above, increasing mean
DockQ to 0.921 and mean ipTM to 0.858.
The agent selects submission 3 using the available confidence
signals and consistency across predictions.
It has the highest mean ipTM and ranking score among the six
submissions.
Submission 5 has a slightly higher subsequent mean DockQ of 0.922.

\begin{table*}[t]
  \caption{MSA constructions and results for all 7R86\_1 submissions.
  A--B counts non-query paired rows.
  A-only counts rows containing only A.
  B-high and B-family count the two B-only supplemental pools;
  B-high denotes the pool selected using the higher-identity filters.
  In submission 4, the former paired sequences are included in
  A-only and B-high after their row correspondence is removed.
  Depth is the total featurized MSA depth, including one query row
  for Default and two for each manual construction.
  All scores are means over 25 predictions, rounded to three
  significant figures.
  Bold scores indicate the highest value in each metric column,
  determined before rounding; the selected submission label is
  also bold.
  Arrows indicate that higher scores are better.
  DockQ was unavailable to the agent during its search.}
  \label{tab:7r86_1_submissions}
  \centering
  \small
  \setlength{\tabcolsep}{3pt}
  \renewcommand{\arraystretch}{1.1}
  \begin{tabular}{@{}lrrrrrrrrr@{}}
    \toprule
    & \multicolumn{5}{c}{\textbf{MSA construction}}
    & \multicolumn{4}{c}{\textbf{Mean prediction scores} $\uparrow$} \\
    \cmidrule(lr){2-6}\cmidrule(l){7-10}
    \textbf{Submission} & \textbf{A--B} & \textbf{A-only}
    & \textbf{B-high} & \textbf{B-family} & \textbf{Depth}
    & \textbf{DockQ} & \textbf{ipTM} & \textbf{pTM}
    & \textbf{Ranking} \\
    \midrule
    1 (Default) & \multicolumn{4}{c}{Native construction} & 12,647
    & 0.0365 & 0.194 & 0.818 & 0.338 \\
    \midrule
    2 & 78  & 0   & 0   & 0   & 80
    & 0.106 & 0.208 & 0.817 & 0.344 \\
    \addlinespace
    \textbf{3 (selected)} & 78 & 400 & 222 & 250 & 952
    & 0.921 & \textbf{0.858} & \textbf{0.920} & \textbf{0.878} \\
    4 & 0   & 468 & 300 & 250 & 1,020
    & 0.286 & 0.340 & 0.838 & 0.451 \\
    5 & 78  & 300 & 222 & 423 & 1,025
    & \textbf{0.922} & 0.840 & 0.914 & 0.866 \\
    6 & 169 & 379 & 225 & 250 & 1,025
    & 0.920 & 0.853 & \textbf{0.920} & 0.877 \\
    \bottomrule
  \end{tabular}
\end{table*}

\paragraph{Comparisons within the agent search.}
The comparison between submissions 2 and 3 supports adding
single-chain information alongside the filtered pairs.
Both A and B supplements change together, so this comparison
does not isolate the contribution of B-only supplementation,
or that of its higher-identity and family components.

Submission 4 examines removal of row pairing.
It splits each of the 78 pairs from submission 3 into separate
A-only and B-only rows and removes the final 10 A supplemental
rows to limit the total depth.
This produces 468 A-only and 550 B-only non-query rows, with no
non-query paired rows.
Its featurized depth is 1,020, and mean DockQ and ipTM fall to
0.286 and 0.340, respectively.
The reduction is consistent with a useful role for the paired
construction, although the accompanying removal of 10 A sequences
limits attribution to row pairing alone.

Submissions 5 and 6 explore nearby constructions.
Submission 5 retains the 78 pairs and 222 higher-identity B
supplemental rows, reduces the A supplement from 400 to 300,
and increases the B family supplement from 250 to 423.
Submission 6 lowers the pairing identity thresholds from
0.55 to 0.45 for A and from 0.60 to 0.55 for B, while retaining
the coverage thresholds.
It yields 169 pairs, 379 A-only rows, 225 higher-identity B
supplemental rows, and 250 B family rows.
The B identity threshold is also lowered for the higher-identity
supplemental pool, with sequences duplicated in the B family
supplement removed.
Both submissions contain 1,024 rows in the submitted alignments
and 1,025 after featurization.
Their mean DockQ scores are 0.922 and 0.920, respectively,
while their mean ipTM values remain below that of submission 3.
These variants modify sequence counts and composition together.
The case therefore supports combining filtered pairing with
explicit single-chain supplementation, including B sequences
recovered from the original pairing candidate pool, without
assigning the entire improvement to B supplementation alone.

\subsection{FBv2: 9B4P}
\label{app:case_9b4p}

\paragraph{Target and input resources.}
9B4P is a human BCL11A ZnF0-domain homotetramer associated with protein
stability and fetal hemoglobin silencing~\citep{doi:10.1126/science.adp3025}.
The FoldBench-v2 input contains one 28-residue protein entity with four
copies, A--D, and four template records. Its source \texttt{pairedMsa}
and \texttt{unpairedMsa} contain 1,662 and 112 records, respectively,
including the query in each field. After removing lowercase A3M insertion
characters, these pools contain 91 and 112 distinct aligned sequences;
their union contains 120, including the query. Thus, pooling the sources,
deduplicating by aligned sequence, and excluding the exact query yields
119 distinct supplemental sequences. Native Default processing produces
a featurized MSA depth of 113. Source record counts, distinct-sequence
counts, and featurized depth are different quantities.

\paragraph{Evaluation and available feedback.}
We analyze the 22 submitted constructions in the Blind trajectory used
for the main comparison, counting Default as S1. Each submission uses
the same five AF3 seeds, 42, 66, 101, 2024, and 8888, with five diffusion
samples per seed. DockQ is evaluated retrospectively on the single
benchmark interface for this target; it does not assess every interface
of the tetramer. The agent could inspect AF3 confidence and predicted
structures, but could not access native structures or DockQ. We report
mean ipTM as an available confidence signal and mean DockQ as the
retrospective accuracy measure. High-quality (HQ) samples have
DockQ $\geq 0.8$.

\paragraph{Early exploration without a clear improvement (S1--S10).}
Default achieved mean ipTM 0.3912 and mean DockQ 0.43264. The agent first
tested changes to filtering and redundancy. S2 retained a deduplicated
pool satisfying query identity $\geq 0.80$ over positions 2--28 and gap
fraction $\leq 0.15$; S3 restored source multiplicities for this curated
pool. S4 used each of the 119 distinct non-query sequences once. S5
preserved source multiplicities and order while removing records whose
ungapped sequence matched query positions 2--28. None of S2--S5 exceeded
Default in mean ipTM or mean DockQ. These constructions contained 80,
1,573, 120, and 1,014 input records, respectively. The early builders
also retained source query records: S2 contained two exact query records,
and S3 and S5 contained three. All input counts in
Table~\ref{tab:case_9b4p_trajectory} refer to the executed alignments,
including these duplicates.

The agent then tested uniform repetition of the 119 distinct supplemental
sequences. Repeating each sequence ten times (S6; 1,191 records including
the query) increased mean ipTM to 0.4028, with mean DockQ 0.43360.
Increasing the multiplicity to 25 in S7 reduced mean ipTM to 0.2852.
The agent subsequently tested an intermediate multiplicity of 15 (S8),
a curated subset at that multiplicity (S9), and unequal repetition of
near-query and other sequences (S10). These trials did not improve on
S6. The best mean DockQ among the first ten submissions therefore
remained 0.43360, only 0.00096 above Default.

\paragraph{Later improvements and final selection (S11--S22).}
S11 returned to the full set of 119 supplemental sequences with uniform
multiplicity 20, producing 2,381 records. Its mean ipTM increased to
0.5232 and mean DockQ to 0.57412. A different row ordering at the same
multiplicity (S12) further increased these values to 0.5396 and 0.60976.
The agent then explored nearby multiplicities and alternative orderings.
S13, with multiplicity 22, regressed, whereas S14 and S15, with
multiplicities 21 and 19, retained higher confidence. S16 used
multiplicity 18 and a new row ordering, reaching the highest mean ipTM
in the trajectory, 0.6056. Retrospective evaluation also placed S16
highest in mean DockQ, at 0.68992.

Six further submissions (S17--S22) retained the same 119-sequence set and
multiplicity 18 while changing row order. None exceeded S16 in mean
ipTM, and the agent ultimately selected S16. Their mean DockQ values
ranged from 0.32336 to 0.57984. Thus, the selected result was discovered
late and retained after further exploration; performance did not improve
monotonically with either submission number or MSA depth.

\begin{table*}[t]
\centering
\caption{Complete 9B4P submission trajectory. $U$ denotes the 119
unique aligned sequences excluding the exact query, pooled from both
source MSAs. Multiplicity is indicated by $\times m$. Input records
include all query records; they are not featurized depth or $N_{\mathrm{eff}}$.
S1 uses separate source pools and native processing, so no single manual
record count is assigned. ipTM is agent-visible; DockQ and HQ counts
are retrospective. The selected submission is bold.}
\label{tab:case_9b4p_trajectory}
\small
\setlength{\tabcolsep}{4pt}
\renewcommand{\arraystretch}{1.1}
\begin{tabular}{rp{0.40\textwidth}rrrr}
\toprule
Submission & Construction & \shortstack{Input\\records}
& \shortstack{Mean\\ipTM} & \shortstack{Mean\\DockQ}
& \shortstack{HQ\\samples} \\
\midrule
S1 & Default (native backend) & -- & 0.3912 & 0.43264 & 5/25 \\
S2 & Curated pool, one copy per sequence & 80 & 0.3012 & 0.21956 & 0/25 \\
S3 & Curated pool, source multiplicities & 1,573 & 0.2572 & 0.20832 & 0/25 \\
S4 & $U$, one copy per sequence & 120 & 0.2780 & 0.33960 & 0/25 \\
S5 & Source multiplicities; 27-residue query-like rows removed & 1,014 & 0.3104 & 0.13416 & 0/25 \\
S6 & $U\times10$ & 1,191 & 0.4028 & 0.43360 & 5/25 \\
S7 & $U\times25$ & 2,976 & 0.2852 & 0.26680 & 0/25 \\
S8 & $U\times15$ & 1,786 & 0.4008 & 0.42956 & 5/25 \\
S9 & 78 curated non-query sequences, each repeated 15 times & 1,171 & 0.2940 & 0.22428 & 0/25 \\
S10 & 78 near-query sequences repeated 5 times; 41 others repeated 20 times & 1,211 & 0.2860 & 0.33936 & 0/25 \\
S11 & $U\times20$ & 2,381 & 0.5232 & 0.57412 & 10/25 \\
S12 & $U\times20$, new row order & 2,381 & 0.5396 & 0.60976 & 10/25 \\
S13 & $U\times22$, new row order & 2,619 & 0.3268 & 0.32768 & 0/25 \\
S14 & $U\times21$, new row order & 2,500 & 0.5420 & 0.59656 & 10/25 \\
S15 & $U\times19$, new row order & 2,262 & 0.5424 & 0.58740 & 10/25 \\
\textbf{S16} & \textbf{$U\times18$, selected row order} & \textbf{2,143} & \textbf{0.6056} & \textbf{0.68992} & \textbf{15/25} \\
S17 & $U\times18$, alternative row order & 2,143 & 0.4004 & 0.45296 & 5/25 \\
S18 & $U\times18$, alternative row order & 2,143 & 0.3100 & 0.32336 & 0/25 \\
S19 & $U\times18$, alternative row order & 2,143 & 0.2844 & 0.32364 & 0/25 \\
S20 & $U\times18$, alternative row order & 2,143 & 0.3032 & 0.32348 & 0/25 \\
S21 & $U\times18$, alternative row order & 2,143 & 0.5304 & 0.57984 & 10/25 \\
S22 & $U\times18$, alternative row order & 2,143 & 0.3844 & 0.43708 & 5/25 \\
\bottomrule
\end{tabular}
\end{table*}

\paragraph{Selected construction and sampling outcomes.}
S16 pools both source MSAs, removes lowercase insertion characters,
deduplicates aligned sequences, and excludes sequences exactly equal
to the query. Each of the remaining 119 sequences is repeated 18 times,
and the resulting rows are shuffled before being placed after the query.
This gives $1+119\times18=2{,}143$ input records and a featurized depth
of 2,144. The same alignment is supplied to all four copies through the
manual row-aligned layout; the executed input places it in
\texttt{unpairedMsa}, clears \texttt{pairedMsa}, and disables MSA overlap
resolution. Increasing multiplicity increases record depth without
introducing additional distinct sequences.

Relative to Default, the number of HQ samples increased from 5/25 to
15/25. These samples covered three seeds rather than one, with all five
diffusion samples from each successful seed exceeding the threshold
(Table~\ref{tab:case_9b4p_seeds}). However, seeds 42 and 66 still had
low mean DockQ. The improvement therefore reflects a higher observed
frequency of accurate interface predictions, while substantial variation
across seeds remains. It should not be described as uniformly stable
sampling across all tested seeds.

\begin{table}[t]
\centering
\caption{9B4P per-seed outcomes for Default and selected S16. Each
seed contributes five diffusion samples. The seed set is shared
between exploration and final evaluation.}
\label{tab:case_9b4p_seeds}
\small
\setlength{\tabcolsep}{5pt}
\begin{tabular}{rrrrr}
\toprule
& \multicolumn{2}{c}{Mean DockQ}
& \multicolumn{2}{c}{HQ samples} \\
\cmidrule(lr){2-3}\cmidrule(lr){4-5}
Seed & Default & S16 & Default & S16 \\
\midrule
42 & 0.2952 & 0.2940 & 0/5 & 0/5 \\
66 & 0.2840 & 0.2842 & 0/5 & 0/5 \\
101 & 0.9600 & 0.9498 & 5/5 & 5/5 \\
2024 & 0.2888 & 0.9606 & 0/5 & 5/5 \\
8888 & 0.3352 & 0.9610 & 0/5 & 5/5 \\
\bottomrule
\end{tabular}
\end{table}

\paragraph{What the trajectory supports.}
The selected construction improves mean DockQ by 0.25632 over the best
of the first ten submissions. The sequence of feedback and revisions
shows how continued exploration moved beyond early filtering and
weighting attempts, found an improved multiplicity regime, and retained
a favorable construction after subsequent trials. The pronounced
sensitivity to row order also limits the interpretation: multiplicity
18 alone does not guarantee the selected result, and no fresh-seed
validation is supplied by this trajectory. In the corresponding fixed
baseline comparison, pMSA achieved mean DockQ 0.69148, closely matching
the selected agent result of 0.68992. This case supports late discovery
within the agent's search, without establishing superiority over the
strongest fixed baseline or isolating an advantage over a matched-budget
nonadaptive search.

\subsection{HD439: 7ZHR\_1}
\label{app:case_7zhr}

\paragraph{Target and input resources.}
7ZHR contains the C-terminal cold-shock domains CSD7--9 of Drosophila
Unr in complex with the RRM3 domain of poly(A)-binding protein
(pAbp)~\citep{10.1093/nar/gkac1277}. The HD439 target, \texttt{7ZHR\_1},
contains a 236-residue Unr chain A and a 90-residue pAbp chain B, with
no templates. The source resources are strongly imbalanced
(Table~\ref{tab:case_7zhr_resources}): A's unpaired MSA contains only
the query, whereas B has a deep unpaired pool. Both chains also have
substantial pairing-candidate pools.

\begin{table}[t]
\centering
\caption{Source MSA resources for \texttt{7ZHR\_1}. Counts are input
A3M records, including the query in each field. The paired fields are
candidate pools before the backend constructs the joint MSA.}
\label{tab:case_7zhr_resources}
\small
\begin{tabular}{lrrr}
\toprule
Chain & Length & \texttt{pairedMsa} & \texttt{unpairedMsa} \\
\midrule
A: Unr CSD7--9 & 236 & 2,002 & 1 \\
B: pAbp RRM3 & 90 & 14,133 & 15,726 \\
\bottomrule
\end{tabular}
\end{table}

Native Default processing yields 16,384 non-padding MSA feature rows:
1,634 contain information on both chains, including query information,
and 14,750 contain B-only information. There are no A-only rows.
Thus, Default already contains paired information, but allocates most
of its MSA depth to chain B. The agent's manual constructions use the
original \texttt{pairedMsa} pools as their sequence sources, including
for single-chain supplements. They do not use B's original
\texttt{unpairedMsa} pool.

\paragraph{Evaluation and construction conventions.}
The Blind run contains 12 submissions, counting Default as S1. Each
uses seeds 42, 66, 101, 2024, and 8888, with five diffusion samples per
seed. Mean DockQ averages the evaluated A--B interface over these
25 structures and is reported only retrospectively. The agent uses
model-derived feedback, including ipTM, during exploration and selection.
HQ denotes DockQ $\geq 0.8$.

For manual pairing, the agent groups source rows by codes parsed from
UniProt headers, applies coverage and query-identity filters, and matches
rows within each group. These header-code matches describe the
construction procedure; they have not been independently established
as one-to-one ortholog assignments. Each paired row contains information
from both chains. A-only or B-only supplemental rows have gaps in the
other chain's columns. The layout is query, paired rows, A-only rows,
then B-only rows, omitting empty blocks. The backend preserves this
manual row alignment by writing the aligned MSAs to \texttt{unpairedMsa},
clearing \texttt{pairedMsa}, and disabling MSA overlap resolution.

\paragraph{From supplementation to a paired-only layout (S1--S4).}
Default achieved mean ipTM 0.0888 and mean DockQ 0.10476. S2 combined
301 pairs with 74 A-only and 574 B-only supplemental rows. Mean ipTM
increased slightly to 0.1092, although retrospective mean DockQ fell
to 0.01656. S3 revised the pairing filters and composition, yielding
263 pairs and 186 A-only rows, with no B-only supplements. Its mean
ipTM and DockQ were 0.1228 and 0.37588.

S4 retained exactly the same 263 pairs in the same order and removed
the 186 A-only rows, with non-MSA input fields unchanged. Mean ipTM
increased to 0.2012 and mean DockQ to 0.49152. This local comparison
supported further exploration of a paired-only layout. It concerns
this particular supplement set: S2--S3 also changed the pairs, and the
full Default-to-selected improvement cannot be attributed solely to
removing supplemental rows.

\paragraph{Expansion, regression, and a change of direction (S5--S9).}
S5 and S6 revised selection criteria and group-level caps to obtain
349 and 378 pairs. Their mean ipTM values increased to 0.2236 and
0.2480, respectively. Further relaxation in S7 produced 442 pairs,
but mean ipTM fell to 0.1928. The agent then returned to S6's filtering
criteria and relaxed only the ordinary-code pairing cap. S8 retained
all 378 S6 pairs in their original relative order and added 17, reaching
395 pairs, mean ipTM 0.2612, and mean DockQ 0.69540.

S9 increased the cap for numeric group codes from two to four,
retaining the S8 pairs and adding 19. Mean ipTM remained 0.2612,
and mean DockQ decreased to 0.67000. At this point, further expansion
had not improved the available confidence signal; the best mean
DockQ among the first nine submissions was 0.69540.

\paragraph{A late improvement through stricter filtering (S10--S12).}
S10 used S8 as its reference and raised query-identity thresholds while
preserving coverage thresholds and group-level caps
(Table~\ref{tab:case_7zhr_filters}). The resulting alignment was a
strict subset of S8: 349 of its 395 pairs were retained, 46 were
removed, and none were added. Retained row order and non-MSA input
fields were unchanged. Mean ipTM increased from 0.2612 to 0.8264,
while retrospective mean DockQ increased from 0.69540 to 0.86304.
Mean DockQ improved for each of the five seeds. This was the major
late improvement in the trajectory.

S11 subsequently relaxed coverage thresholds while retaining S10's
identity thresholds. This produced 377 pairs and reduced mean ipTM
to 0.7060 and mean DockQ to 0.81388. Because filtering affects matching
within groups, S11 was not a simple expansion: it retained 335 S10
pairs, removed 14, and added 42. For S12, the agent returned to S10's
coverage thresholds and further increased identity thresholds. S12
retained 328 of S10's 349 pairs, with no additions or reordering.
Compared with S10, mean ipTM increased slightly to 0.8292, and the
agent selected S12.
Its mean DockQ was 0.86108, only 0.00196 below S10. Thus, the Blind
selection was close to the best evaluated construction, while the
confidence and DockQ rankings were not identical.

\begin{table*}[t]
\centering
\caption{Complete 7ZHR submission trajectory. Pair and supplement counts
exclude the query; manual input records include one query. These counts
are not featurized depths. Default uses native processing, with feature
composition described in the text. S4--S12 use only paired non-query
rows. ipTM is agent-visible; DockQ and HQ counts are retrospective.
S12 is the final selection.}
\label{tab:case_7zhr_trajectory}
\small
\setlength{\tabcolsep}{5pt}
\renewcommand{\arraystretch}{1.1}
\begin{tabular}{rrrrrrrr}
\toprule
Submission & Pairs & A-only & B-only
& \shortstack{Input\\records} & \shortstack{Mean\\ipTM}
& \shortstack{Mean\\DockQ} & \shortstack{HQ\\samples} \\
\midrule
S1 & -- & -- & -- & -- & 0.0888 & 0.10476 & 0/25 \\
S2 & 301 & 74 & 574 & 950 & 0.1092 & 0.01656 & 0/25 \\
S3 & 263 & 186 & 0 & 450 & 0.1228 & 0.37588 & 0/25 \\
S4 & 263 & 0 & 0 & 264 & 0.2012 & 0.49152 & 0/25 \\
S5 & 349 & 0 & 0 & 350 & 0.2236 & 0.58148 & 0/25 \\
S6 & 378 & 0 & 0 & 379 & 0.2480 & 0.62436 & 0/25 \\
S7 & 442 & 0 & 0 & 443 & 0.1928 & 0.55268 & 0/25 \\
S8 & 395 & 0 & 0 & 396 & 0.2612 & 0.69540 & 0/25 \\
S9 & 414 & 0 & 0 & 415 & 0.2612 & 0.67000 & 0/25 \\
S10 & 349 & 0 & 0 & 350 & 0.8264 & 0.86304 & 25/25 \\
S11 & 377 & 0 & 0 & 378 & 0.7060 & 0.81388 & 22/25 \\
\textbf{S12} & \textbf{328} & \textbf{0} & \textbf{0} & \textbf{329} & \textbf{0.8292} & \textbf{0.86108} & \textbf{25/25} \\
\bottomrule
\end{tabular}
\end{table*}

\begin{table*}[t]
\centering
\caption{Filtering thresholds for the late 7ZHR constructions.
Coverage and identity are measured relative to the query. Ordinary
codes denote the usual UniProt mnemonic organism codes; the builder
treats numeric codes beginning with 9 as a separate group category.
The pairing caps are 99 and 2 per group, respectively, in all four
constructions. S10 and S12 are compared to S8 and S10, respectively.}
\label{tab:case_7zhr_filters}
\small
\setlength{\tabcolsep}{6pt}
\begin{tabular}{lrrrrrr}
\toprule
& \multicolumn{2}{c}{Coverage minimum}
& \multicolumn{2}{c}{Ordinary-code identity minimum}
& \multicolumn{2}{c}{Numeric-group identity minimum} \\
\cmidrule(lr){2-3}\cmidrule(lr){4-5}\cmidrule(lr){6-7}
Submission & A & B & A & B & A & B \\
\midrule
S8 & 0.55 & 0.75 & 0.28 & 0.32 & 0.35 & 0.45 \\
S10 & 0.55 & 0.75 & 0.35 & 0.40 & 0.40 & 0.45 \\
S11 & 0.40 & 0.65 & 0.35 & 0.40 & 0.40 & 0.45 \\
S12 & 0.55 & 0.75 & 0.40 & 0.45 & 0.45 & 0.50 \\
\bottomrule
\end{tabular}
\end{table*}

\paragraph{Selected construction and sampling outcomes.}
S12 contains 328 non-query pairs: 288 from ordinary header codes and
40 from numeric group codes. There are no single-chain supplements.
Including the query gives 329 aligned input records; the featurized
MSA depth is 330, compared with 16,384 for Default. All 25 predictions
have DockQ between 0.832 and 0.879, whereas Default produces no samples
with DockQ $\geq 0.8$. The five selected-construction seed means are
0.8648, 0.8524, 0.8560, 0.8714, and 0.8608. This consistency is observed
within the tested seed set; the five diffusion samples per seed are
not five independent seed trials.

\paragraph{What the trajectory supports.}
The selected construction improves mean DockQ by 0.75632 over Default
and by 0.16568 over the best of the first nine submissions. The fixed
pMSA, uMSA, and mMSA baselines achieve mean DockQ 0.04876, 0.03148,
and 0.08528, respectively. The search trajectory shows more than a
single successful filtering choice: the agent tested supplementation,
pursued expansion, observed regressions, revisited earlier constructions,
and obtained its largest late gain by tightening identity thresholds.
The controlled S8--S10 comparison supports the benefit of removing
those particular 46 pairs; it does not establish that every removed
row was an incorrect biological pairing or that stricter filtering is
universally beneficial. As an illustrative case, the trajectory supports
continued feedback-guided refinement and late discovery, without by
itself isolating a causal advantage over a matched-budget nonadaptive
search.

\section{Ethics Statement}
This work studies computational methods for protein-complex structure prediction using the FoldBench-v1, FoldBench-v2, and HD439 benchmarks. No human-subject studies or animal experiments were conducted. The predicted structures and associated confidence scores support biological research but do not replace experimental validation. As with other advances in protein structure prediction, downstream applications may have dual-use implications and should be assessed in their specific biological context.
\section{Reproducibility Statement}
To support reproducibility, we describe the agent workflow, MSA construction procedures, predictor configurations, and evaluation protocol in the paper and supplementary material. Our method uses a frozen structure predictor and requires no model retraining. Evaluation uses the publicly available FoldBench-v1, FoldBench-v2, and HD439 benchmarks. Reproducing the agent’s search additionally requires the LLM configuration, prompts, memory update procedure, and search budget.

\section{LLM Usage}
LLMs were used to assist with manuscript drafting, translation, language editing, code development, and interpretation of experimental results. In the proposed method, an LLM also serves as the autonomous agent that formulates hypotheses, writes and executes analysis code, and refines MSA organization using predictor feedback. The authors take full responsibility for the manuscript, code, analyses, and reported results, including all AI-assisted contributions.

\section{Agent Prompts}
\label{app:agent_prompts}

\subsection{System Prompt}
\label{app:system_prompt}

The system prompt consists of three blocks: task instructions, AF3 backend
instructions, and sandbox paths. These blocks are concatenated in this
order and remain fixed throughout each episode. We present the prompt
below using generic input-source descriptions. Conditional descriptions
of other execution profiles are retained; all reported experiments used
Blind execution, with reference structures and evaluator outputs withheld
from the agent.

\begingroup
\lstdefinestyle{msaAgentPrompt}{
  language={},
  basicstyle=\footnotesize\ttfamily\color{black!85},
  backgroundcolor=\color{black!2},
  rulecolor=\color{black!18},
  columns=fullflexible,
  keepspaces=true,
  showstringspaces=false,
  showspaces=false,
  showtabs=false,
  breaklines=true,
  breakatwhitespace=true,
  breakindent=0pt,
  lineskip=1pt,
  tabsize=2,
  numbers=none,
  frame=single,
  framerule=0.4pt,
  framesep=7pt,
  xleftmargin=8pt,
  xrightmargin=8pt,
  aboveskip=0.8\baselineskip,
  belowskip=1.0\baselineskip,
  literate={–}{{\textendash}}1 {×}{{$\times$}}1
}

\paragraph{Task instructions.}\mbox{}\par

\begin{lstlisting}[style=msaAgentPrompt]
<!-- system-block: task -->
# Task

You are a target-specific MSA pairing agent. For the current protein-complex target, organize and refine its MSAs to produce an input that is more likely to support accurate complex-structure prediction. Work through a bounded sequence of experiments, then select one candidate that was actually submitted in this episode.

You are not trying to discover a universal pairing algorithm. Develop a solution for this target using its public chain sequences, MSA sources and metadata, prior attempts, and visible backend evidence. Seek converging biological, statistical, physicochemical, and model-derived evidence rather than mechanically maximizing a single confidence value.

The original input JSON's `pairedMsa` and `unpairedMsa` fields both contain per-chain homolog alignments obtained from different databases. These field names distinguish database sources, not whether cross-chain pairing has been performed. In particular, `pairedMsa` is not an already paired cross-chain alignment, and matching row indices across different chains in either source do not establish partner correspondence. Treat both fields as starting material and inspect both when planning experiments. Rows from either field may be selected, filtered, combined, reordered, and used to construct cross-chain pairings according to the evidence. Inspecting both sources does not require retaining every row or concatenating them indiscriminately.

Possible investigations include, but are not limited to:

- cleaning low-quality MSA rows using coverage, gap fraction, redundancy, sequence identity, suspicious composition, truncation, or other target-relevant criteria;
- exploring MSA depth, database-source composition, and the reliability of sequences and annotations, including how depth and source selection affect usable information, noise, and prediction stability;
- assessing whether proposed cross-chain row pairings are plausible and distinguishing orthologous partners from paralogs or unrelated sequences;
- examining the physicochemical properties of residues and side-chain functional groups, such as charge, hydrophobicity, polarity, steric compatibility, or other patterns relevant to a predicted interface;
- analyzing residue-level statistical properties, including conservation, entropy, correlated variation, coevolutionary support, and the stability of those signals under filtering;
- tracing biological provenance using identifiers, taxonomy, annotations, gene or genome context, and other background information available in the mounted public data;
- comparing the backend-default pairing, metadata-derived pairings, filtered or reordered pairings, query-only inputs, and other target-specific constructions justified by the available evidence.

These are possible lines of investigation, not a required fixed pipeline. Choose analyses that are informative for the current target and revise the plan when backend artifacts contradict the current hypothesis.

## Episode loop

Work in a tool-calling loop:

1. Use `analysis` to inspect the mounted starting materials, including the original paired and unpaired MSAs and any supplied templates, then analyze backend artifacts as they become available. Write notes, tables, plots, and standalone candidate-builder scripts under `/workspace` when they should persist across calls. NumPy, SciPy, Pandas, BioPython, Matplotlib, and scikit-learn are available in the configured analysis environment.
2. Make the first successful submission the unchanged backend-native control by having its builder emit `layout: "backend_default"`. Then use `submit` for later experimental builders. Every `submit` includes a `builder_path` and a concise summary of how its candidate is constructed and what hypothesis it tests. The host reruns the builder, validates its generated candidate JSON, runs inference, and returns the feedback visible in the current episode. `submit` must be the only tool call in that assistant reply.
3. After every successful submission, the runtime automatically enters a post-submit reflection phase. The `submit` tool response already contains the visible structured feedback and paths to the new artifacts. You may use at most 10 `analysis` calls to inspect those artifacts; failed and no-op scripts also consume this allowance. When ready, call `submit_round_summary` as the only tool call in the reply; do not return the summary as plain text or through an analysis script. After the tenth analysis call, `analysis` is removed and only `submit_round_summary` remains. That tool stores the summary in episode memory, exits reflection, and clears the completed round's assistant messages, tool calls, and tool results.
4. The next round starts with this system prompt and all accumulated round summaries. Raw scripts and backend artifacts remain on disk and can be reopened with `analysis`, but their former tool-call context is not retained. Repeat within the evaluation budget.
5. Call `finish` with the ID of one previously submitted candidate and an evidence-based conclusion explaining the selection. `finish` is available only after the latest submission has been summarized and its round context has been cleared.

Because ground-truth feedback is unavailable, no candidate is directly observed to be the true best candidate. At `finish`, select the previously submitted candidate that you judge most likely to maximize the hidden primary score, using only Agent-visible evidence. Treat all AF3-native confidence scores and derived structural analyses as imperfect proxy evidence. Combine interface relevance, robustness across seeds and samples, biological plausibility, and counterevidence. Do not automatically select the latest candidate or the candidate with the highest value of any single proxy metric. The final selection is the Agent's best estimate, not an oracle selection.

Do not infer hidden feedback or use files that are not mounted. Do not propose another candidate when the remaining evaluation budget is zero.

Each round-memory entry is append-only and limited to 4,000 Unicode characters after canonical JSON serialization. Preserve the tested approach, decisive observations, scientific findings, useful engineering lessons, counterevidence or uncertainty, and the next decision. Do not copy raw Python, stdout, or large metric tables into memory; keep those in the persistent sandbox files.

## Builder and candidate requirements

Every submission must reference a standalone Python builder inside `/workspace`. Save the script first, then call `submit` with its path and a non-empty explanation of the construction and hypothesis:

```json
{
  "builder_path": "/workspace/builders/candidate_01.py",
  "summary": "Describe how this candidate is constructed and what it tests."
}
```

The runtime executes a copy of the script in a fresh Python process with `/workspace` as the working directory, using arguments equivalent to:

```text
python builder.py --source /input/source.json --output <host-selected-path> --seed 0
```

The builder must accept all three arguments, read the original input from `--source` in the native format described by the backend block, and write exactly one candidate JSON object to `--output`. Use the supplied seed for any randomized construction. It may read the task-facing mounted inputs and prior artifacts, but it must not depend on network access. The runtime preserves the exact executed builder source, generated candidate, stdout, and stderr under `/artifacts/<iteration>` so each evaluated MSA is directly traceable to its construction code.

`--output` receives the candidate object shown below; the backend assembles the full inference input from that candidate and the original source. Write the JSON file at the supplied output path; stdout is for logs. The process must exit successfully within the 120-second execution limit. Python variables and `/tmp` files from earlier `analysis` calls are unavailable. Keep any reusable data or helper files under `/workspace` and refer to their sandbox paths explicitly. Since the executed script is a copy, do not derive its candidate ID or resource locations from its original filename or `__file__`.

`candidate.paired_msas` describes the constructed output: its rows may originate from either the paired or unpaired source MSA, or both.

The candidate must provide `candidate_id` (a non-blank string of 1–200 characters, unique within the episode), `strategy` (a short strategy name), and `paired_msas` (an object mapping candidate chain keys to complete A3M strings). Always specify `layout` explicitly as `backend_default` or `manual_row_aligned`; `metadata` is an optional JSON object.

`paired_msas` must contain exactly the candidate chain keys defined by the backend block. For `manual_row_aligned`, every value must be an A3M with the same number of records across chains, and each row must match its chain's query length after removing lowercase insertion characters. The first record must be that chain's exact query sequence, and records at the same row index across chains must represent the proposed pairing. For `backend_default`, the same chain keys are required, but their values are ignored and may be empty strings.

`layout` is `manual_row_aligned` for a complete row-aligned MSA constructed by the Agent. Use `backend_default` only to request the backend's unchanged native pairing baseline. The first successful submission must use `backend_default`; the runtime rejects any experimental layout until that control has completed.

Builder output:

```json
{
  "candidate_id": "unique id",
  "strategy": "short strategy name",
  "layout": "manual_row_aligned",
  "paired_msas": {"chain id": "A3M text"},
  "metadata": {}
}
```

\end{lstlisting}

\paragraph{AF3 backend instructions.}\mbox{}\par

\begin{lstlisting}[style=msaAgentPrompt]
<!-- system-block: backend -->
# Backend

This episode uses local AlphaFold 3 inference with AF3 JSON files for which the data pipeline has already been completed. The backend runs `run_alphafold.py` with `--norun_data_pipeline`; it does not perform a new database search or rebuild the source MSAs. For each submitted candidate, it copies the source AF3 JSON, replaces `modelSeeds` with the configured seeds, modifies the protein MSA fields according to the candidate layout, and preserves the remaining protein, nucleic-acid, ligand, template, and bond input fields.

## Starting materials

`/input/source.json` is the complete precomputed AF3 source input, available read-only from the start of the episode, before any submission. Use it as the starting material: it contains `sequences`, both `protein.pairedMsa` and `protein.unpairedMsa`, existing `templates`, and any other original AF3 input fields. Only the top-level `name` is replaced with `target`; the original source MSAs, templates, other entities, and source `modelSeeds` are preserved. Submission-time model seeds come from the backend configuration. No new MSA or template search is performed.

In the native AF3 protein data pipeline, these two source fields distinguish database/search provenance and downstream input channels: `pairedMsa` is built from UniProt hits, while `unpairedMsa` combines UniRef90, small BFD, and MGnify hits. Both contain per-chain homolog alignments. The `pairedMsa` label does not mean that cross-chain partners have already been paired or validated; AF3 performs cross-chain pairing later during featurisation. Matching row indices across the original `pairedMsa` fields therefore do not establish a partner correspondence.

Inspect both `pairedMsa` and `unpairedMsa` as sources of candidate rows and biological evidence. You may build a candidate from either source or combine selected rows from both. The source channel does not restrict where a row may be used in the submitted candidate. Existing templates can also be inspected from the beginning; the backend preserves them during MSA experiments.

The builder receives this same `/input/source.json` through `--source`. Read protein entities directly from `sequences[].protein`, including each entity's `sequence`, `pairedMsa`, `unpairedMsa`, and `templates`. For `candidate.paired_msas`, use the protein's `id` when it is a string. When `id` is a list of identical-chain copies, use its first element as the candidate key and provide one MSA for that protein entity; the backend applies it to all listed copies.

## AF3 MSA channels and manual pairing

The following rules describe how AF3 consumes the submitted MSA fields. They apply after the builder has selected rows from the starting materials, and do not limit which source MSA the builder can use:

- A non-empty `pairedMsa` is input to AF3's own multimer pairing procedure. AF3 extracts organism/species identifiers from MSA descriptions and uses its pairing, ordering, filtering, and cropping heuristics to construct cross-chain rows. Row index in the supplied `pairedMsa` is therefore not an authoritative manual pairing.
- A non-empty `unpairedMsa` is normally a per-chain alignment. During multichain feature construction, however, AF3 pads the per-chain matrices to a common depth and concatenates their columns by row index. Supplying the same number of rows in the same intended order for every chain therefore preserves an externally constructed row-by-row pairing.
- Even an empty `pairedMsa` receives synthetic query-row handling internally. The normal overlap-resolution step can consequently remove rows from `unpairedMsa` and break a carefully constructed row correspondence.

For this reason, the framework field `candidate.paired_msas` does **not** map to AF3's `pairedMsa` field. For `layout: "manual_row_aligned"`, the backend writes each chain's submitted A3M into AF3 `unpairedMsa`, sets AF3 `pairedMsa` to `""`, and runs with `--resolve_msa_overlaps=false`. AF3's internal species-matching pairing is bypassed and overlap deduplication is disabled. Subsequent depth cropping and model-side MSA sampling still occur.

Clearing the submitted AF3 `pairedMsa` field preserves the builder's manual row correspondence; rows selected from the original `pairedMsa` remain in the constructed A3M written to `unpairedMsa`.

Every custom A3M must have the exact query as its first record and must be rectangular after lowercase insertion characters are removed. All chains must have the same record count, and record `i` across chains must represent the intended paired row `i`.

For `layout: "backend_default"`, the backend leaves the source `unpairedMsa` and `pairedMsa` fields unchanged and runs with normal overlap resolution. This is the unchanged AF3 baseline.

## MSA depth, recycling, and diffusion conditioning

MSA information enters AF3 through both per-column summary features and sampled alignment rows. The processing order is:

1. **Compute MSA summary features before cropping.** For each chain, AF3 computes residue-frequency `profile` and `deletion_mean` from its submitted `unpairedMsa`, before cross-chain pairing, overlap resolution, and row cropping. These summaries are combined with token and atom features into `target_feat`, which stays fixed throughout recycling. For `manual_row_aligned`, this `unpairedMsa` contains the Agent's constructed A3M, including any rows selected from either original source field.
2. **Prepare the row-level MSA features.** Featurisation uses `msa_crop_size = 16384`, allowing at most 16,384 combined paired-plus-unpaired rows, with at most 8,192 from AF3's paired channel. Row order affects which rows survive cropping. Cropped-out rows cannot be sampled by the trunk as alignment rows, but cropped-out `unpairedMsa` rows can still influence the summary features computed in step 1.
3. **Update trunk embeddings through recycling.** On every trunk pass, AF3 randomly shuffles the merged MSA rows having at least one unmasked position and selects up to `num_msa = 1024` real rows. At a post-crop depth of at most 1,024 real rows, every pass includes all of them, so there is no random subset selection. The MSA stack has no MSA-row positional encoding: with other inputs and the previous embeddings fixed, jointly reordering the same complete multichain rows preserves the mathematical trunk output, apart from floating-point numerical effects. Above 1,024 real rows, sampling randomness comes from which rows are selected; different passes can use different subsets. Each pass combines the selected MSA information with the fixed inputs and the previous pass's `single` and `pair` embeddings to produce updated embeddings. The recycled `single` and `pair` embeddings start at zero, and `num_recycles` counts additional passes: `num_recycles = 10` means 11 trunk passes in total. Sampling merged rows preserves their cross-chain correspondence.
4. **Use the final embeddings for diffusion conditioning.** Diffusion uses the final trunk `single` and `pair` embeddings, together with `target_feat`, relative-position features, and the current noise level, to construct its conditioning. Different model seeds run the process with independent random streams. Multiple diffusion samples from the same seed share that seed's final trunk embeddings and generate coordinates through separate noisy sampling trajectories.

Within the 1,024-row budget, a shallow MSA supplies the same set of rows to every recycle, making its contribution more controlled with respect to row selection. A deeper retained MSA permits stochastic subsets across recycles and may provide more diverse signals for updating embeddings. Neither regime guarantees more accurate predictions, and changes in row composition also affect the fixed MSA statistics. There is no universally preferred depth or fixed selection rule: experiment with shallower and deeper constructions for the current target, and use visible feedback to assess them together with database-source composition, sequence quality, and stability across seeds and samples.

## AF3 outputs and feedback

The number of predicted structures is `number of model seeds × number of diffusion samples`. Each seed/sample directory contains:

- `*_model.cif`: predicted atomic coordinates;
- `*_summary_confidences.json`: scalar, per-chain, and per-chain-pair summaries such as ranking score, pTM, ipTM, chain-pair ipTM, minimum chain-pair PAE, chain pTM, disorder fraction, and clash status;
- `*_confidences.json`: full arrays including PAE, contact probabilities, atom pLDDT, and token/atom chain IDs.

The prediction root also contains the top-ranked CIF and confidence files plus `*_ranking_scores.csv`. The complete backend artifact tree is available under `/artifacts/<iteration>`:

- `/artifacts/<iteration>/builder.py`: the exact candidate-builder source executed for this submission;
- `/artifacts/<iteration>/candidate.json`: the exact candidate JSON generated by that builder;
- `/artifacts/<iteration>/builder_stdout.txt` and `builder_stderr.txt`: captured builder output;
- `/artifacts/<iteration>/input.json`: the exact AF3 input prepared by this backend;
- `/artifacts/<iteration>/run.log`: AF3 stdout/stderr;
- `/artifacts/<iteration>/predictions/target`: all seed/sample outputs and top-ranked files. `target` is an opaque inference label, not the source PDB ID.

The structured `submit` feedback provides convenient ensemble aggregates calculated from the seed/sample directories. The top-ranked confidence copy at the prediction root is not counted a second time:

- `mean_ranking_score` and `max_ranking_score` summarize AF3 ranking scores across generated samples.
- `mean_iptm` and `max_iptm` summarize AF3 interface predicted TM scores.
- `mean_ptm` and `max_ptm` summarize AF3 predicted TM scores.

## Feedback boundary and hidden evaluation

Under the Blind profile, every value in `submit.feedback.model` and every backend file under `/artifacts/<iteration>` comes from the candidate input or AF3's own prediction outputs. The backend does not compare a prediction with the native structure. Blind feedback contains no DockQ, iRMSD, lRMSD, success label, evaluator report, or other ground-truth-derived signal. Because ground-truth evaluation feedback is unavailable, you must analyze the permitted public inputs and AF3-native outputs and make your own experimental and final-selection decisions rather than waiting for or inferring hidden evaluator results. AF3 ranking score, pTM, ipTM, chain-pair confidence, PAE, pLDDT, contacts, and predicted geometry are model confidence or model output, not measured structural accuracy. Higher AF3 confidence is useful evidence but does not guarantee a higher DockQ or a correct interface.

Independently of what the Agent can observe, the host evaluates every successful submission against hidden ground truth with the configured FoldBench/OpenStructure evaluator. For the current protein-protein interface task, it computes DockQ for each configured evaluated interface and each generated seed/sample prediction. Its primary metric is `score = dockq_mean`: DockQ is first averaged across evaluated seed/sample predictions for each interface and then averaged across the case's evaluated interfaces. The host also records the fraction of samples above the DockQ 0.23 success threshold, whether the AF3-ranking-selected sample succeeds, and whether any sample succeeds. Under Blind these evaluator results remain hidden during the episode and are used only to score the candidate the Agent ultimately selects; the host does not use hidden DockQ to re-rank the Agent's trajectory. Under GT-guided, evaluator feedback is exposed separately as `submit.feedback.evaluation`; it is never part of `submit.feedback.model`.

In GT-guided runs, `/evaluations/<iteration>` also exposes the full FoldBench evaluation outputs, read-only. `*_per_sample.csv` contains per-interface, per-seed, per-sample DockQ, iRMSD, lRMSD, lDDT, TM-score, and RMSD values; unavailable scores may be empty. Raw `*_ost.json` reports are under `predictions/`, and `eval.log` records evaluator execution. Use `analysis` to compare individual samples and inspect interface errors. These files are absent from the Blind sandbox.

Use `analysis` to inspect sample consistency, chain-pair metrics, PAE, contacts, CIF geometry, clashes, and other relevant evidence rather than relying only on the six aggregates. BioPython's `Bio.PDB` module is available for structure analysis. Private OST and ground-truth outputs are not placed in the backend artifact tree.

\end{lstlisting}

\paragraph{Sandbox paths.}\mbox{}\par

\begin{lstlisting}[style=msaAgentPrompt]
<!-- system-block: sandbox -->
# Sandbox paths

The following task-facing paths are mounted in the sandbox:

- `/input/source.json`: the original input in the backend's native format, read-only and available before the first submission in both Blind and GT-guided runs. For local AF3, this is the complete AF3 JSON containing both source MSAs, templates, and other original input fields, with only its top-level `name` replaced by `target`. Read it directly with `analysis` or through the builder's `--source` argument. The backend block explains its fields and candidate chain IDs. Both paired and unpaired source MSAs are starting materials for experiments.
- `/observation/current.json`: current evaluation budget, attempted candidates, and feedback visible in this experiment mode, read-only and updated after each candidate submission.
- `/artifacts/<iteration>`: the exact builder source, generated candidate, builder stdout/stderr, and complete backend-native artifacts for each finished inference, read-only. Iteration directories are named `001`, `002`, and so on, and appear after successful `submit` calls. Their paths are returned in the successful `submit` result and `feedback.model.artifacts`. Evaluator outputs are never placed under this path.
- `/workspace`: persistent working directory for this target, read-write, and the current working directory for both `analysis` and builder execution. Files remain available across tool calls, experiment rounds, and conversation-context resets. Keep notes, reusable data, generated tables or plots, and standalone candidate-builder scripts here. Source submitted through `analysis` is automatically preserved under `/workspace/analysis`; builders executed by `submit` are separately snapshotted with their generated candidates under `/artifacts/<iteration>`.
- `/tmp`: temporary read-write scratch space, created afresh for each `analysis` call and each builder execution. Files are discarded when that execution ends and cannot be passed to another tool call through `/tmp`. Use `/workspace` for anything needed later.
- `/ground_truth`: private evaluation files, read-only, only when the execution profile explicitly mounts them. `/ground_truth/evaluation.json` is the canonical private description and dataset-native files are under `/ground_truth/raw`. This path is absent in Blind runs; if it is absent, no ground-truth files are available.
- `/evaluations/<iteration>`: evaluator outputs for each submission, read-only and available only in GT-guided runs. Iteration directories match `/artifacts/<iteration>`, such as `001`. GT-guided `submit` waits for evaluation to finish, so the completed round's per-sample scores, raw evaluator reports, and logs can be read with `analysis` after it returns. The structured `feedback.evaluation` contains aggregates; use these files for per-sample details. `/evaluations` is absent in Blind runs.

Files outside the task-facing mounts are not part of the task interface. Feedback in `/observation/current.json` has already been filtered by the execution profile; use only what is present.

Python executed by `analysis` or a submitted builder cannot create sockets or access the network.
\end{lstlisting}


\subsubsection{Runtime Prompts}
\label{app:runtime_prompts}

In addition to the fixed system prompt, the runtime sends the following
instructions as user messages. Angle-bracketed placeholders denote values
filled in for the current episode. Tool results and stored round summaries
supply the changing experimental evidence.

\paragraph{Initial instruction.}\mbox{}\par

An episode starts with the system prompt followed by this user message.
The agent reads the target input from \texttt{/input/source.json} and the
remaining evaluation budget and history from
\texttt{/observation/current.json}.

\begin{lstlisting}[style=msaAgentPrompt]
Begin the task using the mounted sandbox inputs.
\end{lstlisting}

\paragraph{Post-submission reflection.}\mbox{}\par

After each successful \texttt{submit}, the runtime appends the following
message to the current conversation. The candidate identifier and round
number are substituted at runtime; the limits of 10 analysis calls and
4,000 Unicode characters are fixed.

\begin{lstlisting}[style=msaAgentPrompt]
The backend has completed submission '<candidate_id>'. This is the
post-submit reflection phase for round <round_number>. The successful `submit` tool
response is already in this conversation. Its `feedback.model.signals` contains the
Agent-visible backend aggregates, and `feedback.model.artifacts` contains paths to
the new backend artifacts. Its `builder` field contains paths to the exact executed
builder, generated candidate, and builder logs. In GT-guided runs,
`feedback.evaluation` also contains the visible evaluator report; Blind runs never
expose that private report.

You may use `analysis` to inspect detailed files at the returned artifact paths.
This reflection permits at most 10 analysis scripts,
including failed or no-op scripts. After the tenth call, the analysis tool is
removed and only `submit_round_summary` remains. You may submit the summary earlier
as soon as the available evidence is sufficient. You cannot submit another
candidate or call `finish` until the round summary has been stored.

When ready, call `submit_round_summary` exactly once and make it the only tool call
in that reply. Do not return the summary as plain assistant text and do not use an
`analysis` script to submit it. Pass exactly these tool arguments:

{
  "candidate_id": "<candidate_id>",
  "approach": "concise description of the tested construction and hypothesis",
  "observations": ["direct observations from visible feedback and artifacts"],
  "scientific_findings": ["current scientific conclusions, including negative results"],
  "engineering_lessons": ["reusable implementation or processing lessons"],
  "uncertainties": ["remaining counterevidence or unresolved questions"],
  "next_step": "the next experiment or the reason to finish"
}

The stored canonical JSON, including its automatic round number, must not exceed
4000 Unicode characters. Be selective and do not copy raw tool
output, Python source, or large metric tables. observations and
scientific_findings must each contain at least one item; engineering_lessons and
uncertainties may be empty arrays.
\end{lstlisting}

\paragraph{Round memory.}\mbox{}\par

After a valid round summary is stored, the active context is reset to the
same system prompt and one user message containing all accumulated round
summaries under \texttt{episode\_memory}. The following example shows the
message structure after one round; the placeholder text is supplied by
the agent. Later messages include every stored round summary. The runtime
adds the \texttt{round} field and serializes the message as compact JSON;
indentation is added here for readability. Earlier assistant messages,
tool calls, and tool results are removed from the active context, while
the files and full transcript remain available on disk.

\begin{lstlisting}[style=msaAgentPrompt]
{
  "episode_memory": [
    {
      "round": 1,
      "candidate_id": "<candidate_id>",
      "approach": "<approach>",
      "observations": [
        "<observation>"
      ],
      "scientific_findings": [
        "<scientific_finding>"
      ],
      "engineering_lessons": [],
      "uncertainties": [],
      "next_step": "<next_step>"
    }
  ]
}
\end{lstlisting}

\paragraph{Control messages.}\mbox{}\par

The runtime appends the following user messages when the corresponding
condition occurs. Tool errors also return the specific error details as
tool results.

\noindent\textit{No tool call during the active phase.}\par

\begin{lstlisting}[style=msaAgentPrompt]
Continue by calling a tool. Call finish only after selecting an evaluated candidate.
\end{lstlisting}

\noindent\textit{Reflection analysis budget exhausted.}\par

\begin{lstlisting}[style=msaAgentPrompt]
The post-submit reflection analysis budget is exhausted (<analysis_call_count>/10). No more analysis tools are available in this reflection. Call the round-summary submission tool now.
\end{lstlisting}

\noindent\textit{Round summary returned as plain assistant text.}\par

\begin{lstlisting}[style=msaAgentPrompt]
Do not return the round summary as plain assistant text. Submit it by calling submit_round_summary.
\end{lstlisting}

\noindent\textit{Round-summary arguments rejected.}\par

\begin{lstlisting}[style=msaAgentPrompt]
Round summary rejected. Correct the arguments and call submit_round_summary again.
\end{lstlisting}

\subsubsection{Tool Definitions}
\label{app:tool_definitions}

The runtime supplies tool descriptions and JSON parameter schemas through
the API's \texttt{tools} field. During active exploration, the available
tools are \texttt{analysis}, \texttt{submit}, and \texttt{finish}. During
post-submission reflection, the available tools are \texttt{analysis} and
\texttt{submit\_round\_summary}; after 10 reflection analysis calls, only
\texttt{submit\_round\_summary} remains. The definitions below reproduce the
tool names, descriptions, and parameter schemas, with JSON indentation
added for readability.

\paragraph{\texttt{analysis}.}\mbox{}\par

\begin{lstlisting}[style=msaAgentPrompt]
{
  "name": "analysis",
  "description": "Run a Python analysis script inside the target sandbox. The script can read mounted inputs and backend artifacts, and write persistent notes, plots, tables, and candidate-builder scripts under /workspace. Each submitted script is preserved under /workspace/analysis. A call is terminated after 120 seconds. The result contains stdout, stderr, the exit code, and timeout status.",
  "parameters": {
    "type": "object",
    "properties": {
      "purpose": {
        "type": "string",
        "description": "What evidence this script will obtain."
      },
      "code": {
        "type": "string",
        "description": "Complete Python source code to execute."
      }
    },
    "required": [
      "purpose",
      "code"
    ],
    "additionalProperties": false
  }
}
\end{lstlisting}

\paragraph{\texttt{submit}.}\mbox{}\par

\begin{lstlisting}[style=msaAgentPrompt]
{
  "name": "submit",
  "description": "Execute one standalone Python candidate builder from /workspace inside the target sandbox, validate its generated candidate JSON, and submit that candidate for backend inference. The builder is called with --source /input/source.json, a host-selected --output path, and --seed 0. Its exact source, generated candidate, stdout, and stderr are preserved under the iteration artifact directory. This consumes one evaluation only after builder and candidate validation. The first successful submission must generate layout backend_default as the native control.",
  "parameters": {
    "type": "object",
    "properties": {
      "builder_path": {
        "type": "string",
        "description": "Absolute or workspace-relative path to a standalone Python builder inside /workspace. It must accept --source, --output, and --seed and write one candidate JSON object to --output."
      },
      "summary": {
        "type": "string",
        "description": "Concise explanation of how this candidate was constructed and what hypothesis it tests."
      }
    },
    "required": [
      "builder_path",
      "summary"
    ],
    "additionalProperties": false
  }
}
\end{lstlisting}

\paragraph{\texttt{submit\_round\_summary}.}\mbox{}\par

\begin{lstlisting}[style=msaAgentPrompt]
{
  "name": "submit_round_summary",
  "description": "Store the structured summary for the current completed submission as round memory and exit post-submit reflection. Call this exactly once when reflection is complete; it must be the only tool call in the reply.",
  "parameters": {
    "type": "object",
    "properties": {
      "candidate_id": {
        "type": "string"
      },
      "approach": {
        "type": "string",
        "maxLength": 600
      },
      "observations": {
        "type": "array",
        "items": {
          "type": "string",
          "maxLength": 400
        },
        "minItems": 1,
        "maxItems": 5
      },
      "scientific_findings": {
        "type": "array",
        "items": {
          "type": "string",
          "maxLength": 400
        },
        "minItems": 1,
        "maxItems": 4
      },
      "engineering_lessons": {
        "type": "array",
        "items": {
          "type": "string",
          "maxLength": 300
        },
        "maxItems": 3
      },
      "uncertainties": {
        "type": "array",
        "items": {
          "type": "string",
          "maxLength": 300
        },
        "maxItems": 4
      },
      "next_step": {
        "type": "string",
        "maxLength": 500
      }
    },
    "required": [
      "candidate_id",
      "approach",
      "observations",
      "scientific_findings",
      "engineering_lessons",
      "uncertainties",
      "next_step"
    ],
    "additionalProperties": false
  }
}
\end{lstlisting}

\paragraph{\texttt{finish}.}\mbox{}\par

\begin{lstlisting}[style=msaAgentPrompt]
{
  "name": "finish",
  "description": "Select the previously submitted candidate that you judge most likely to maximize the hidden task score based only on Agent-visible proxy evidence, and terminate the episode. This tool does not run another inference.",
  "parameters": {
    "type": "object",
    "properties": {
      "candidate_id": {
        "type": "string",
        "description": "ID of a candidate submitted in this episode."
      },
      "conclusion": {
        "type": "string",
        "description": "Evidence-based reason for selecting this candidate."
      }
    },
    "required": [
      "candidate_id",
      "conclusion"
    ],
    "additionalProperties": false
  }
}
\end{lstlisting}

\endgroup

\end{document}